\documentclass[sigconf]{acmart}

\usepackage{booktabs} 
\usepackage{algorithm}
\usepackage{algpseudocode}
\usepackage{subfigure}
\usepackage{tikz}
\setcopyright{rightsretained}

\acmDOI{10.475/123_4}

\acmISBN{123-4567-24-567/08/06}

\acmConference[KDD'18]{}{August 19-23 2018}{London, United Kingdom} 
\acmYear{2018}
\copyrightyear{2018}

\acmArticle{4}
\acmPrice{15.00}

\editor{}
\editor{}
\editor{}

\begin{document}
\title{Fast Interactive Image Retrieval using large-scale unlabeled data}
  
\author{Akshay Mehra}
\affiliation{%
  \institution{The Ohio State University}
  \streetaddress{}
  \city{} 
  \state{} 
  \postcode{}
}
\email{mehra.42@osu.edu}

\author{Jihun Hamm}
\affiliation{%
  \institution{The Ohio State University}
  \streetaddress{}
  \city{} 
  \state{} 
  \postcode{}
}
\email{hammj@cse.ohio-state.edu}

\author{Mikhail Belkin}
\affiliation{%
  \institution{The Ohio State University}
  \streetaddress{}
  \city{} 
  \state{} 
  \postcode{}
}
\email{mbelkin@cse.ohio-state.edu}

\renewcommand{\shortauthors}{Mehra et al.}

\begin{abstract}
An interactive image retrieval system learns which images in the database belong to a user's query concept, by analyzing the example images and feedback provided by the user. The challenge is to retrieve the relevant images with minimal user interaction. In this work, we propose to solve this problem by posing it as a binary classification task of classifying all images in the database as being relevant or irrelevant to the user's query concept. Our method combines active learning with graph-based semi-supervised learning (GSSL) to tackle this problem. Active learning reduces the number of user interactions by querying the labels of the most informative points and GSSL allows to use abundant unlabeled data along with the limited labeled data provided by the user. To efficiently find the most informative point, we use an uncertainty sampling based method that queries the label of the point nearest to the decision boundary of the classifier. We estimate this decision boundary using our heuristic of adaptive threshold. To utilize huge volumes of unlabeled data we use an efficient approximation based method that reduces the complexity of GSSL from $O(n^3)$ to $O(n)$, making GSSL scalable. We make the classifier robust to the diversity and noisy labels associated with images in large databases by incorporating information from multiple modalities such as visual information extracted from deep learning based models and semantic information extracted from the WordNet. High F1 scores within few relevance feedback rounds in our experiments with concepts defined on AnimalWithAttributes and Imagenet (1.2 million images) datasets indicate the effectiveness and scalability of our approach.
\end{abstract}


\maketitle

\section{Introduction}
Traditionally, image retrieval systems relied on keyword annotation to search for images relevant to a user's query. Although these systems were effective but they suffered from several problems such as the need to annotate huge image databases, deal with inconsistencies among the annotations provided by different annotators etc. To overcome these problems content-based image retrieval (CBIR) \cite{veltkamp2001content, smeulders2000content} was proposed, which used the low level features (such as color, shape etc.) extracted from the images. However, their performance suffered because the low level image features could not capture the high level semantics of an image. Relevance feedback (RF) techniques \cite{rui1998relevance} were used to bridge this semantic gap. But, traditional RF methods required many rounds of feedback before the system could learn what the user was looking for. Active learning techniques \cite{hoi2009semisupervised, hoi2005semi, tong2001support} solved this problem and reduced the number of feedback rounds by identifying the most informative points and asking the user to label only those. Since a common way for a user to provide feedback to an interactive system is by providing binary labels, that indicate whether an image belongs to the query concept or not, we formulate the problem of finding images relevant to a user's query concept from large databases as a binary classification problem. The challenge is to retrieve relevant images with minimal user interaction. In this work, we address this problem by using a method that combines active learning with graph-based semi-supervised learning. Our method learns the user's concept quickly by querying labels of informative points and is scalable to databases with several million images. 

Since the user starts the search with only a small set of labeled images it's not ideal to use supervised learning methods. This is because supervised learning methods require a large number of training examples to perform well. Moreover, they have no way of using the abundant unlabeled data. Combination of active learning and semi-supervised learning can alleviate these problems. Active learning\cite{settles2010active} expands the training set by querying the labels of the most informative points from the pool of unlabeled data. Semi-supervised learning \cite{zhu2006semi}, allows to utilize the abundant unlabeled data and helps to make predictions which are consistent with the inherent graph structure of the data. Many methods have been proposed which use the combination of active and semi-supervised learning \cite{zhu2003combining, hoi2009semisupervised}. However, due to the high computational cost, these methods are not scalable to large datasets such as Imagenet. This is because the manifold regularization framework \cite{belkin2006manifold} for GSSL first builds the neighborhood graph and then propagates the labels. Although we get a closed form solution using this framework but computing that solution requires to invert a large $n$ x $n$ matrix. This is impractical for large applications as it has a time complexity of $O(n^3)$, where $n$ is the total number of points. To use GSSL on a large scale, we use an efficient approximation based method proposed by Fergus et al. \cite{fergus2009semi} which uses the convergence of eigenvectors of a normalized graph Laplacian matrix to eigenfunctions of Laplace-Beltrami operators and brings down the complexity of GSSL from $O(n^3)$ to $O(n)$. The cost of retraining the classifier with this method is also $O(n)$ which makes it possible to quickly retrain the classifier after every round of active learning and incorporate user feedback. We augment this with an uncertainty sampling based method for active learning \cite{campbell2000query, tong2001support} which queries the labels of the points nearest to the decision boundary as they are hardest to classify and hence the most informative. We propose a heuristic based on adaptive threshold to estimate the decision boundary of GSSL classifier and identify the most informative points.

To make the classifier robust to the diversity of the images as well as to the noisy labels associated with images in huge databases, we propose to combine information from multiple modalities such as visual information extracted from state-of-the-art deep learning models such as Resnet\cite{he2016deep}, Xception \cite{chollet2016xception} etc. and semantic information obtained from the WordNet hierarchy\cite{miller1990introduction} in the GSSL method. To achieve this we construct separate graphs using the different features and combine the individual predictions from these graphs. Using the efficient approximation based GSSL method and our heuristic of adaptive threshold, we can classify and find the most informative images in huge databases (>1 million images) in under 5 seconds using a single core cpu machine with 64 GB memory. We then repeat our active learning method to incorporate the user's feedback. We present experiments on concepts defined on AnimalWithAttributes (AWA) \cite{xian2017zero} dataset with 37 thousand images and the Imagenet\cite{deng2009imagenet} dataset with 1.2 million images. Concepts such as ``\textit{furry animals with black stripes}'', ``\textit{person playing a wind instruments}'' etc.~are used for evaluation. The real power of our method lies in its ability to quickly learn from example images provided by the user. Thus, the main contributions of our work can be summarized as follows.

\begin{itemize}
\item We propose a scalable active semi-supervised learning method that uses active learning and efficient approximation based GSSL to quickly learn a user's query concept from a huge database with minimal user interaction. 
   
   \item We propose a method of adaptive threshold that speeds up the learning of a user's concept by iteratively updating the decision boundary of the GSSL classifier for uncertainty sampling based active learning.
     
   \item We present a method that can integrate information from several heterogeneous domains such as visual information from deep learning models, semantic information from WordNet etc. under the GSSL framework.
\end{itemize}

The rest of the paper is organized as follows. In section 2, we briefly discuss some related work. We describe the GSSL framework in section 3 and the method of integrating multimodal features in section 4. In section 5 we present our active learning method. In section 6 we present the results of our experiments on Imagenet and AWA datasets. Finally we conclude in section 7.

\section{Related Work}
Active learning \cite{settles2010active} has been widely applied to CBIR to improve its performance. The works \cite{gosselin2004retin, gosselin2008active} use active learning to understand the user's query concept based on the examples provided. Our work differs from their work as they use an SVM based method for classification which cannot directly use unlabeled data whereas we employ an efficient semi-supervised learning method that can make use of abundant unlabeled data. The works in \cite{hoi2009semisupervised, zhu2003combining} use semi-supervised learning but their method is not scalable to large datasets like the Imagenet. Moreover, their method of choosing the most informative points is also different when compared to our uncertainty sampling based method. Our method of active learning which adds points nearest to the decision boundary is similar to that of \cite{tong2001support}. But the method of \citeauthor{tong2001support} is for SVM based learner where the decision boundary is known where as for our GSSL method we estimate the decision boundary by using an adaptive threshold. Many approaches to CBIR  consider it a ranking problem. These approaches can be inductive or transductive based on whether they use unlabeled data during training or not. Inductive methods \cite{tieu2004boosting, tong2001support} make use of the limited labeled data to learn a classifier that differentiates relevant and irrelevant images. The decision values from this classifier are then used to rank unlabeled data. Transductive methods, on the other hand, use both labeled and unlabeled data. Manifold ranking \cite{zhou2004learning, zhou2004ranking} is a popular transductive method. Significant improvements have been observed by using a large amount of unlabeled data \cite{wan2007manifold, he2004manifold, xu2011efficient, yuan2006manifold}. Our method is close to transductive methods since we also use unlabeled data to make predictions consistent with the inherent geometrical structure of the data. Since GSSL method implicitly gives a continuous output for labels, we can use it directly to rank the images. 
The manifold ranking algorithm can be seen as an extension of the GSSL methodology \cite{belkin2006manifold, zhu2003semi}. The large computational cost of these methods restrict their application to large scale systems. Several works have proposed to improve the performance of these methods. \cite{wang2012scalable} constructs an approximate kNN graph using multiple random divide-and-conquer approach. \cite{tsang2007large} solves the dual optimization problem of manifold regularization \cite{belkin2006manifold} under sparsity constraint. \cite{zhang2009prototype} uses the Nystrom approximation for computing the affinity matrix. \cite{liu2010large} proposed an anchor graph regularization framework which provides an efficient method to create the graph over all the data points. \cite{fergus2009semi} proposed a method for solving the GSSL by working with only a small number of eigenvectors of the graph Laplacian. Our work extends the work of \citeauthor{fergus2009semi}\cite{fergus2009semi} to use multi-modal data. \citeauthor{guillaumin2010multimodal}\cite{guillaumin2010multimodal} proposed a method to combine multi-modal information utilizing tags associated with images, unlike our approach of using class labels and WordNet hierarchy to extract semantic information.

\section{Graph-based Semi-Supervised Learning Framework}
In this section, we introduce the notation for the semi-supervised learning framework that we use and then discuss the issues which arise when we apply the method in a large scale setting and show how to address them.

\subsection{Notation and formulation}
Assume we have a dataset $X$ with $n$ points of which $l$ points  are labeled i.e. $\{(x_{1}, y_{1}), ..., (x_{l}, y_{l})\}$ and remaining $u = n - l$ points are unlabeled $\{x_{l+1}, ..., x_{n}\}$ where $l \ll u$. The labels, $y_{i} \epsilon \{-1, 1\}$. Using all the $n$ points from $X$, we define a graph, $G(V, E)$ with points in $X$ being the vertices $V$ and $E$ being the set of edges between these vertices. The weights on the edges between these points is captured by a $n$ x $n$ affinity matrix, $W$. An example of the function to compute the weights is the radial basis function(RBF) where $W_{ij} = exp( \frac{-1}{2\sigma^{2}} (x_{i} - x_{j})^{2} )$.
Next, we define the graph Laplacian $L = D - W$,
where $D$ is a diagonal matrix with diagonal elements as row sum of $W$, $D_{ii} = \Sigma_{j} W_{ij}$.
We define the objective function for semi-supervised learning using the formulation presented by \citeauthor{fergus2009semi}.
\begin{displaymath}
J(f)= \Sigma_{i = 1}^{l} \lambda(f(i) - y_{i})^{2} + \frac{1}{2}\Sigma_{i,j = 1}^{l+u}(f_{i} - f_{j})^{2} W_{ij}
\end{displaymath}
The first term in the above equation corresponds to the least square loss, which ensures correctness of the labels. The second term defines the smoothness penalty and ensures that the labels of the points are consistent with the manifold and cluster assumptions \cite{zhou2004learning}.  
Simplifying the equation further, we can write $J(f)$ as,
\begin{displaymath}
J(f) = (f-y)^T\Lambda(f-y) + f^{T}Lf
\end{displaymath}
Here $y$ are the labels, $y_{i} \epsilon \{-1, 1\}$ and $\Lambda_{ii} = \lambda$, for labeled points and $\Lambda_{ii} = 0$ for unlabeled points. The optimal $f^{*}$ is the one that minimizes $J(f)$. We can obtain the minimum by setting the gradient of $J(f)$ to zero. The minimizer $f^{*}$ is the solution to the following equation $(L + \Lambda)f = \Lambda y$, neglecting the trivial solution corresponding to a constant $f$ value. The equation  has a closed form solution but it involves inverting a $n$ x $n$ matrix, which has a cubic complexity $O(n^3)$ and is infeasible when $L$ is large. But, as described in \citeauthor{fergus2009semi}, we can reduce the dimension of the problem by working with only a few eigenvectors of the graph Laplacian. Since eigenvectors corresponding to the smallest eigenvalues are the smoothest ( $\Phi_i^TL\Phi_i = \sigma_i$, where $\Phi_i$, $\sigma_i$ are generalized eigenvectors and eigenvalues of L, respectively), hence keeping $k$ eigenvectors corresponding to smallest eigenvalues reduces the dimension of the problem from $n$ x $n$ to $k$ x $k$ assuming we can write $f$ as, $f = U \alpha$ where $U$ is a $k$ x $k$ matrix whose columns are the $k$ eigenvectors corresponding to smallest eigenvalues. Substituting $f=U\alpha$ in $J(f)$ and simplifying, our objective function becomes $J(\alpha) = \alpha^{T}\Sigma\alpha + (U\alpha - y)^{T}\Lambda(U\alpha - y)$
We can now calculate $\alpha$ by solving the following $k$ x $k$ system of equations
\begin{equation} 
(\Sigma + U^{T}\Lambda U)\alpha = U^{T}\Lambda y 
\end{equation}

\subsection{Approximating eigenvectors of $L$}
Even though we can obtain $\alpha$ in equation 1, by solving the $k$ x $k$ system, computing the eigenvectors of the $n$ x $n$ matrix is not easy, as it requires diagonalizing the $n$ x $n$ matrix. Following the analysis presented by \citeauthor{fergus2009semi} which shows the convergence of eigenfunctions as the limit of eigenvectors, as the number of points go to infinity, we compute the eigenfunctions to approximate the eigenvectors of a normalized graph Laplacian. Assuming the data $x_{i} \epsilon R^{d}$ come from a distribution $p(x)$ and rotations make the data independent such that $s = Rx$ then $p(s) = p(s_1)p(s_2)...p(s_d)$. This allows us to calculate the eigenfunctions using only the marginals $p(s_i)$. For each independent component, we can approximate the density $p(s_i)$ using a histogram with $B$ bins. Let $g$ be the eigenfunction values at $B$ discrete points, then $g$ satisfies 
\begin{equation}
(\tilde{D} - P\tilde{W}P)g = \sigma P\hat{D}g
\end{equation}
where $\tilde{W}$ is the affinity between the $B$ discrete points, $P$ is a diagonal matrix whose diagonal elements give the density at the discrete points, and $\tilde{D}$ is a diagonal matrix whose diagonal elements are the sum of the columns of $P \tilde{W} P$, $\hat{D}$ is a diagonal matrix whose diagonal elements are the sum of the columns of $P\tilde{W}$ and $g$ is the eigenfunction values at discrete points.

Following the assumption that data density is dimension separable, we can compute the eigenfunctions for all axes in our data and then keep $k$ eigenfunctions corresponding to the smallest eigenvalues over all the axes. Then we can linearly interpolate the data in 1D and repeat it for all the $k$ eigenfunctions. This step has the complexity of $O(nk)$. Using these $k$ eigenfunctions, which are the approximation to the eigenvectors of the  normalized graph Laplacian, we can solve equation 1 efficiently, since we only need to invert a $k$ x $k$ matrix. 

\begin{algorithm}
\caption{Fergus' Algorithm}
\label{alg:alg-fergus}
\begin{enumerate}
\item (Offline steps)
  \begin{algorithmic}[1]
  \item Perform Principal Component Analaysis (PCA) on input data
  \item For each dimension
      \begin{algorithmic}[1]
      \item Construct a 1D histogram of the marginal
      \item Solve for eigenfunctions and eigenvalues numerically using equation 2
      \end{algorithmic}
  \item Order eigenfunctions from all dimensions by increasing eigenvalues and take first k
  \item Interpolate data using these k eigenfunctions
  \end{algorithmic}
\item (Online steps)
	\begin{algorithmic}[1]
    	\item Update $y$ and $\Lambda$ for labeled points 
    	\item Solve k x k least squares system in equation 1 to obtain the label function $f^*$
    \end{algorithmic}
\end{enumerate}
\end{algorithm}
\vspace{-0.08in}
\section{Integrating Multi-Modal Features}
Here we show how to extend the GSSL framework described in the previous section to incorporate information from multiple modalities. In particular, we show a way to integrate visual and semantic information in the GSSL framework by computing separate graphs for each of them and then combining the individual predictions. 

\subsection{For visual features}
In cases when constructing the exact affinity matrix $\tilde{W}$ is not feasible algorithm \ref{alg:alg-fergus} should be used. We use this method to incorporate visual information from the images. We first extract visual features, for every image, using state-of-the-art deep learning models. After we have these features we use algorithm \ref{alg:alg-fergus} to compute the label function $f^{*}_{visual}$.

\subsection{For semantic features}
Semantic information is often associated with a group or a category of images rather than a single image. This limits the number of unique entries for a large dataset. For example, the number of unique classes in the Imagenet dataset are only 1000 which is significantly less when compared to the number of images in the dataset, 1.2 million. The presence of small number of categories, makes the computation of the exact affinity matrix $\tilde{W}$, in equation 2, possible without the histogram, hence we can simplify algorithm \ref{alg:alg-fergus}. The modified algorithm is presented in algorithm \ref{alg:alg-fergus-modified}. We use algorithm \ref{alg:alg-fergus-modified} to compute the label function for the semantic features associated with the class labels. We use the WordNet hierarchy, Lin similarity \cite{lin1998information} measure and RBF function to compute the affinity matrix $\tilde{W}$ that captures the similarity between the class labels of different images. The formulation used for Lin similarity is defined as in \cite{deselaers2011visual}
\( D^{Lin}(A,B)= \frac{2 log(p(lso(A, B )))}{(log(p(A)) + log(p(B))} \)
where p(A) is the percentage of all images in A, lso(A, B) is the lowest common ancestor of A and B. For example, images with class label ``\textit{cat}'' are closer to the images with class label ``\textit{dog}'' than to images with class label ``\textit{aircraft}''. We compute the label function $f^{*}_{semantic}$ using this information.
\vspace{-0.01in}
\begin{algorithm}
\caption{Modified Fergus' algorithm for semantic features}
\label{alg:alg-fergus-modified}
\begin{enumerate}
\item (Offline steps)
  \begin{algorithmic}[1]
  \item Compute the exact affinity matrix $\tilde{W}$ for all the classes. This matrix reflects the similarity value between different classes.
  \item Using this affinity matrix, solve numerically for eigenfunctions and eigenvalues using equation 2.
  \item Order all the eigenfunctions by increasing eigenvalues and take first k. (Note: there is only a single dimension here)
  \item Construct a matrix of size $Number of Classes$ x $k$, where each row can be treated as a vector associated with one of the classes.
  \item Assign the same vector to each data point belonging to the same class. (This is like the interpolation step) 
  \end{algorithmic}
\item (Online steps)
  \begin{algorithmic}[1]
  \item Update $y$ and $\Lambda$ for labeled points
  \item Solve $k$ x $k$ least squares system in equation 1 to obtain the label function $f^*$
  \end{algorithmic}
\end{enumerate}
\end{algorithm}
\vspace{-0.1in}
\subsection{Combining the predictions}
Once we have the label functions from all the different modalities we obtain the final label function as a convex combination of the individual label functions \(f^{*} = \Sigma_{i = 1}^{k} \lambda_{i}f_{i}^{*}\), where \(\Sigma_{i = 1}^{k}\lambda_{i} = 1\). The parameter $\lambda_{i}$ decides the importance of each modality. This value is dependent on the concept user is trying to search. In our work, we combine the visual information and semantic information by giving equal importance to both. We use $\lambda_{i} = 0.5$ obtain the final label function, $f^{*}$. However, for concepts which purely depend on visual features the value for $\lambda_{visual}$ should be higher than the value of $\lambda_{semantic}$ and it should be the other way for concepts which only rely on semantic information. For example, the concept ``\textit{red objects}'' is a visual concept since color is usually captured by visual features where as ``\textit{works of Picasso}'' is a semantic concept as a user is interested in seeing the works done by Picasso rather than simply seeing works that are visually similar to the provided examples. 

\begin{figure}
\centering
\noindent\subfigure[Mean label values in bin]{\label{fig:makeup-1}\includegraphics[width=0.2\textwidth, height=0.13\textheight]{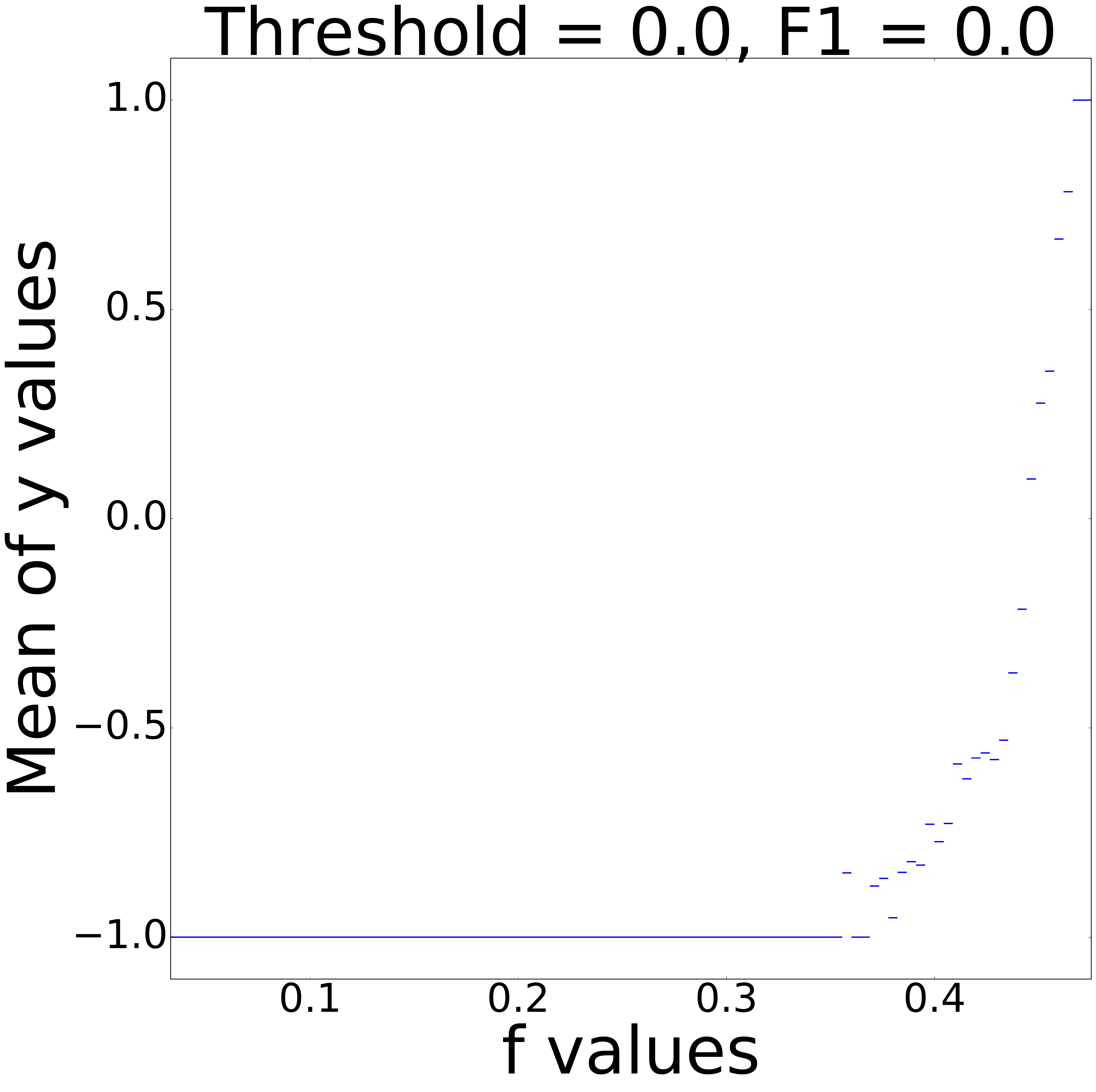}}
    \quad
\noindent\subfigure[Cumulative distribution of $f^*$]{\label{fig:makeup-2}\includegraphics[width=0.2\textwidth, height=0.13\textheight]{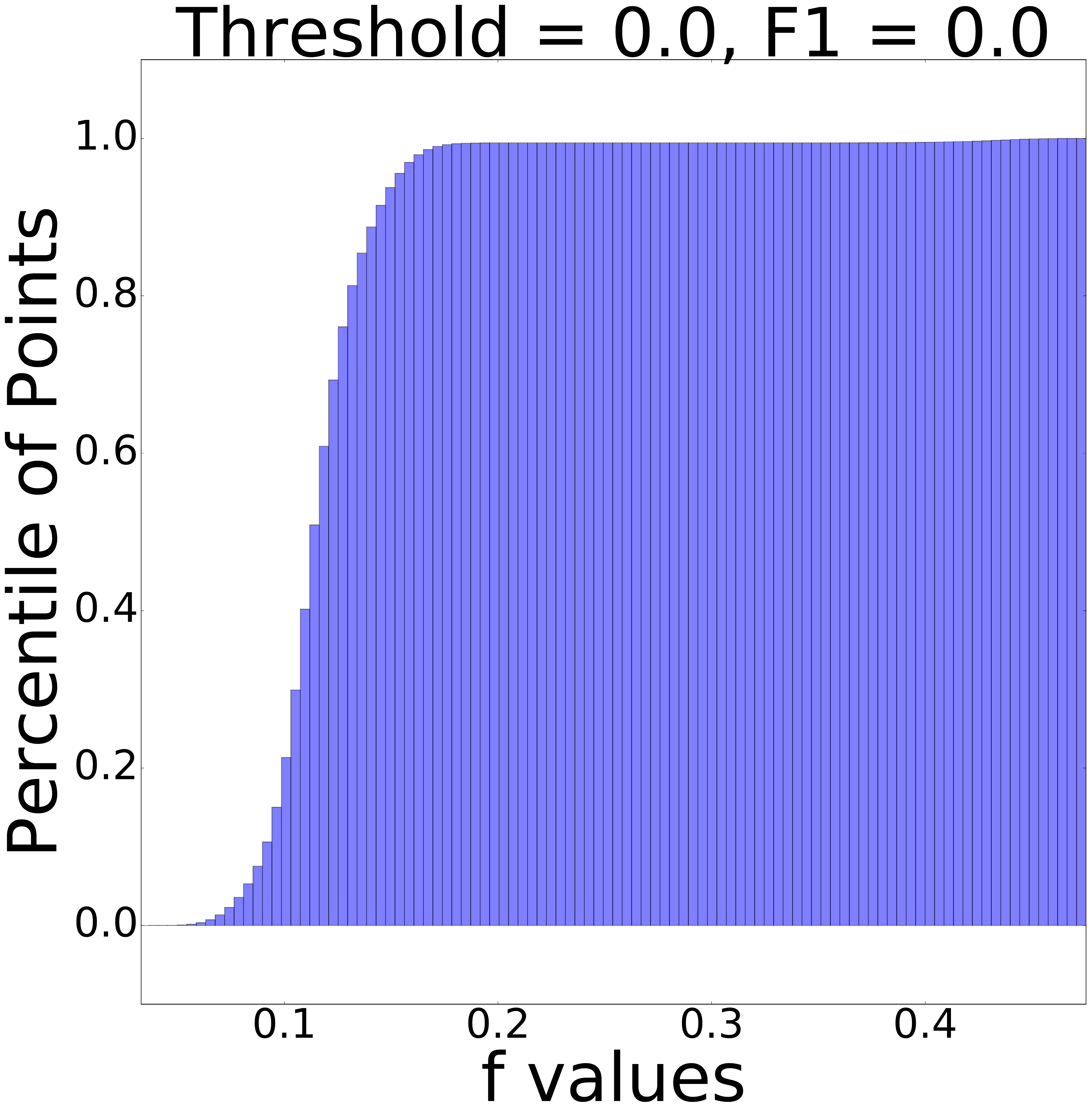}}

\caption{Plot of mean label values for the points in a bin, obtained by creating a histogram of $f^*$ values and cumulative distribution for the $f^*$ values, for the concept ``\textit{makeup accessories}'' with 10 labeled points.}
\label{fig:makeup}
\end{figure}

\section{Active learning for concept detection}
An interactive system learns a user's query concept based on the few example images provided by him. However, these images may not be sufficient for the classifier to understand the concept user is trying to search. Active learning helps to systematically increase the labeled training data by querying the user for labels of the most informative points. Since active learning involves interaction with the user, the aim is to maximize the learning with as few user interactions as possible. In this work, we use an uncertainty sampling based active learning technique which considers the labels of the most ambiguous points to be most informative. We propose a method to find the most ambiguous points in GSSL framework,  which first estimates the decision boundary and then queries the user for labels of the points around it. 

\subsection{Uncertainty sampling}
As shown by \citeauthor{tong2001support}, points around the decision boundary are the most informative ones. However, unlike SVM where the separating hyperplane is known, the decision boundary for the GSSL based method is not known. As seen in the previous section the label function $f^*$ is a continuous output. Since the $f^*$ values are not $p(y|X)$ we must find the decision boundary in a different way. The correct decision boundary is the one that can separate all the positive points from the negative points and can also reflect the class prior correctly. Since the value of $f^*$ for a point is consistent with the manifold and cluster assumptions, we know that a point which is surrounded by more positive points will likely be positive and points surrounded by more negative points will likely be negative. This means the points that have a high $f^*$ would be positive and points with smaller $f^*$ will be negative. Thus, to find the decision boundary we must find a threshold between the range of $f^*$ such that all points above this threshold are positive and all points below this threshold are negative. Once we find such a threshold we can find the points closest to it and ask the user to label them since these are the points which the classifier is most uncertain about.

\begin{figure*}[!ht]
   \centering
  \noindent\subfigure[Zeroth iteration]{\label{fig:makeup-1}\includegraphics[width=0.18\textwidth, height=0.09\textheight]{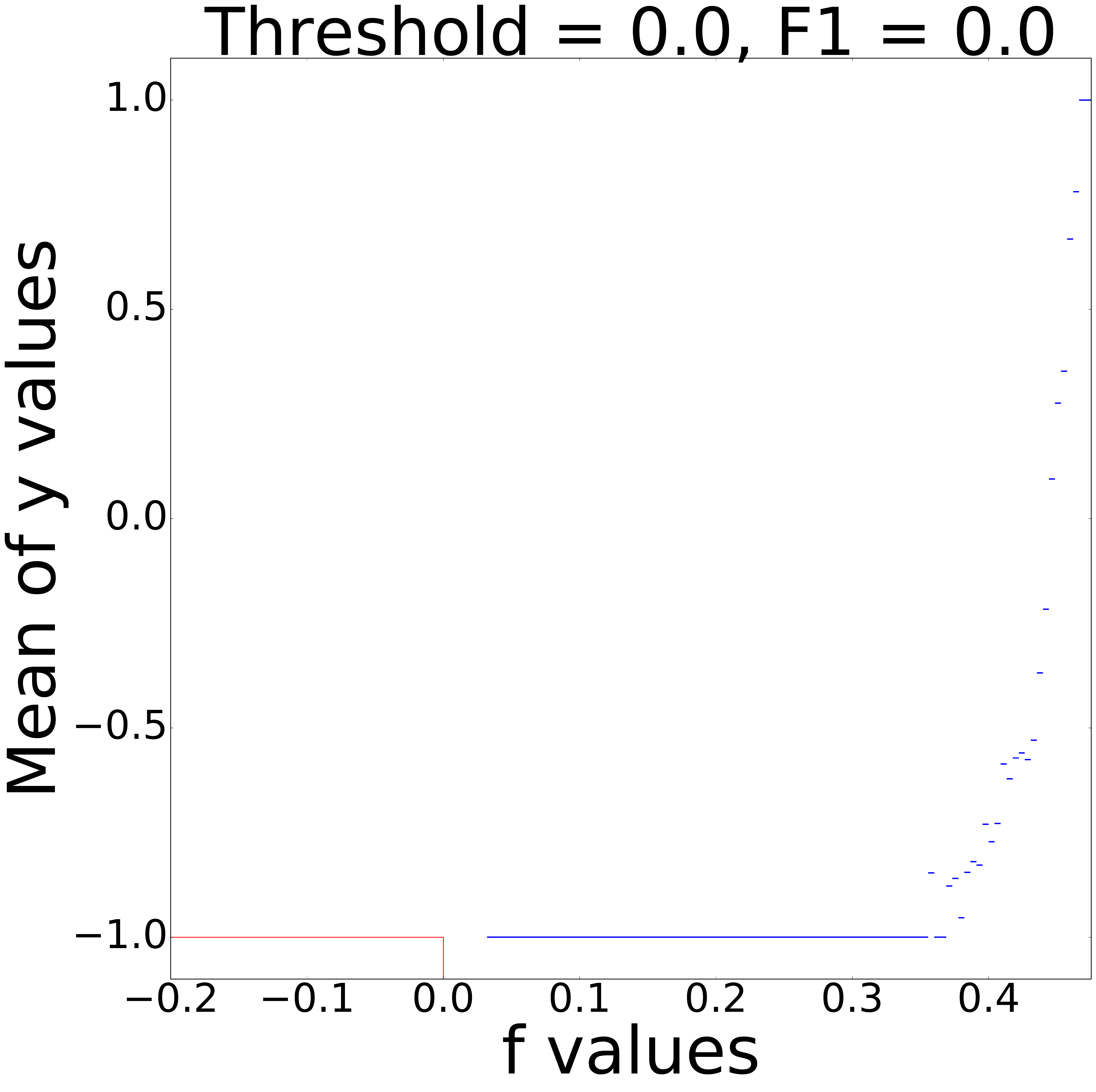}}
  \quad
  \noindent\subfigure[With adaptive threshold after 20 and 200 labeling rounds]{\label{fig:makeup-1}\includegraphics[width=0.18\textwidth, height=0.09\textheight]{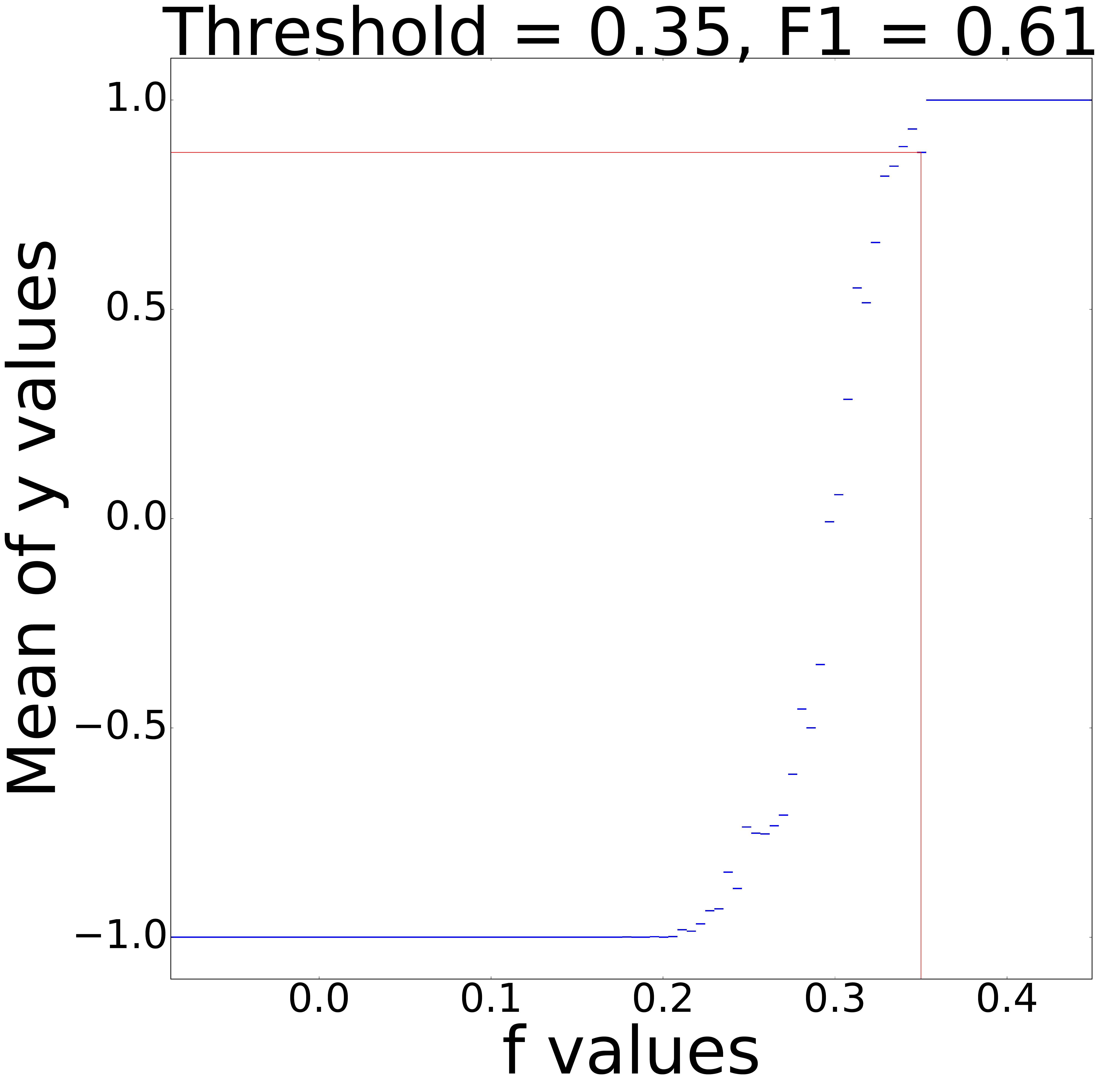}\includegraphics[width=0.18\textwidth, height=0.09\textheight]{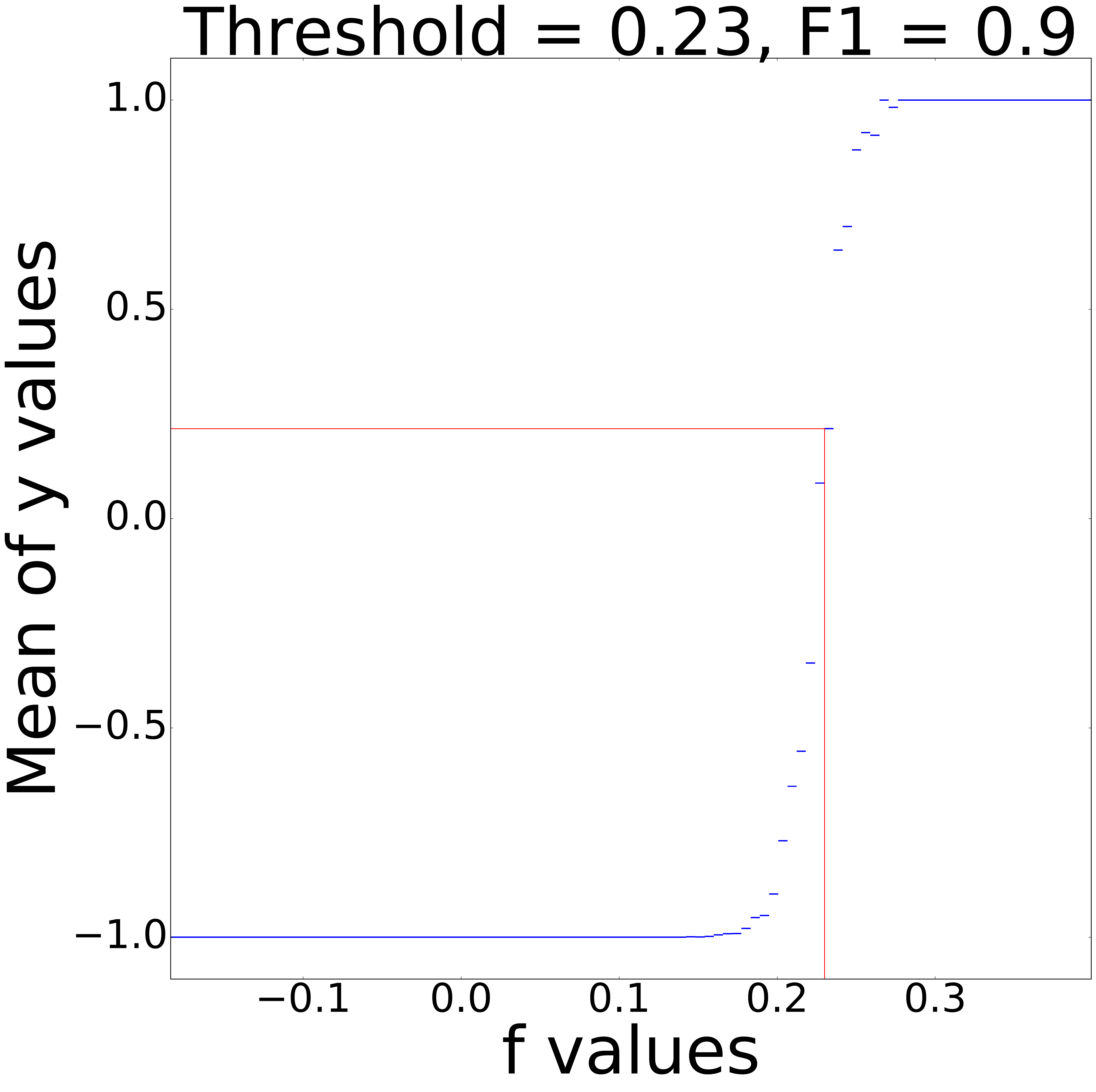}}
  \quad
  \noindent\subfigure[With constant threshold after 20 and 200 labeling rounds]{\label{fig:makeup-1}\includegraphics[width=0.18\textwidth, height=0.09\textheight]{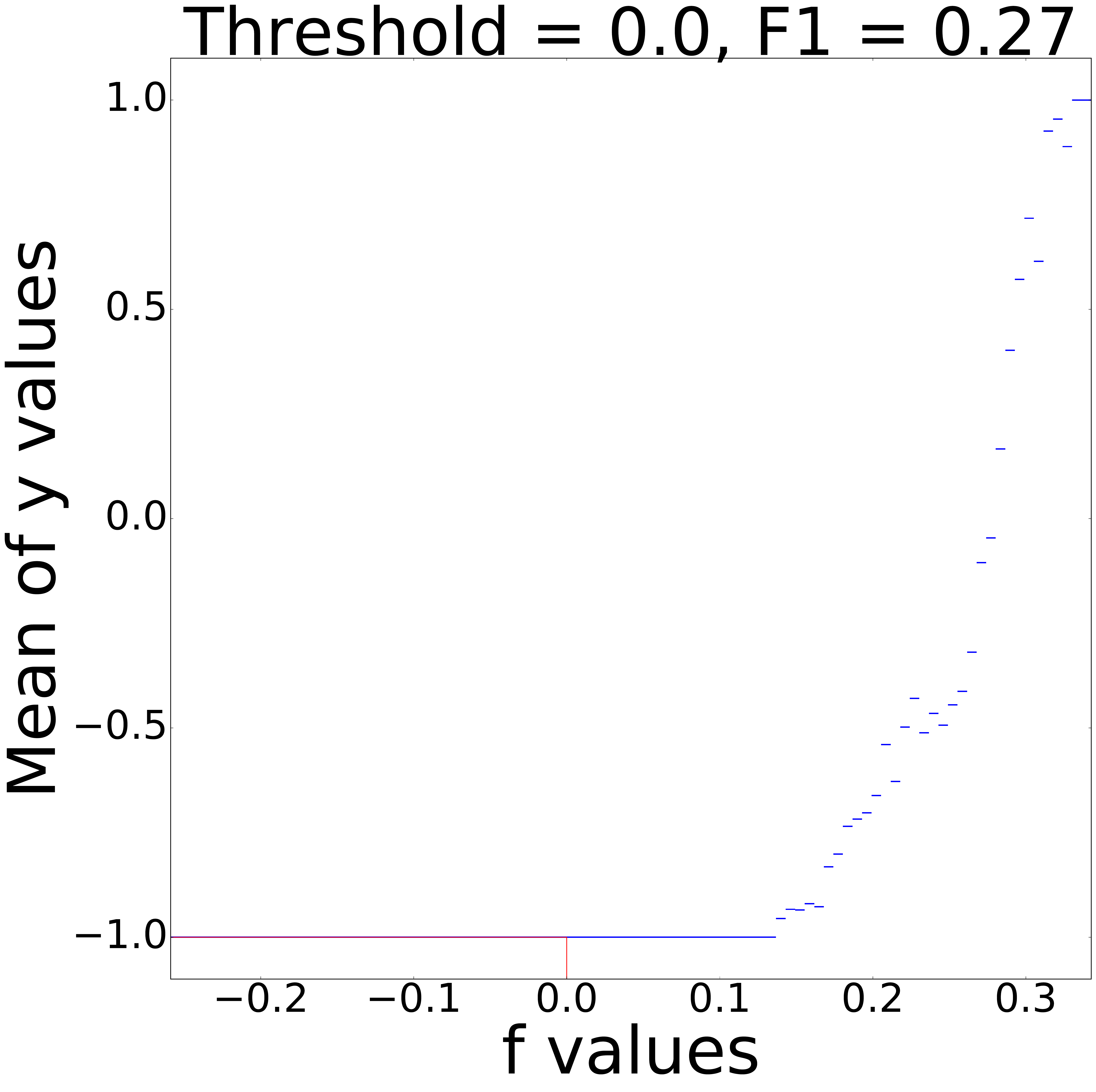}\includegraphics[width=0.18\textwidth, height=0.09\textheight]{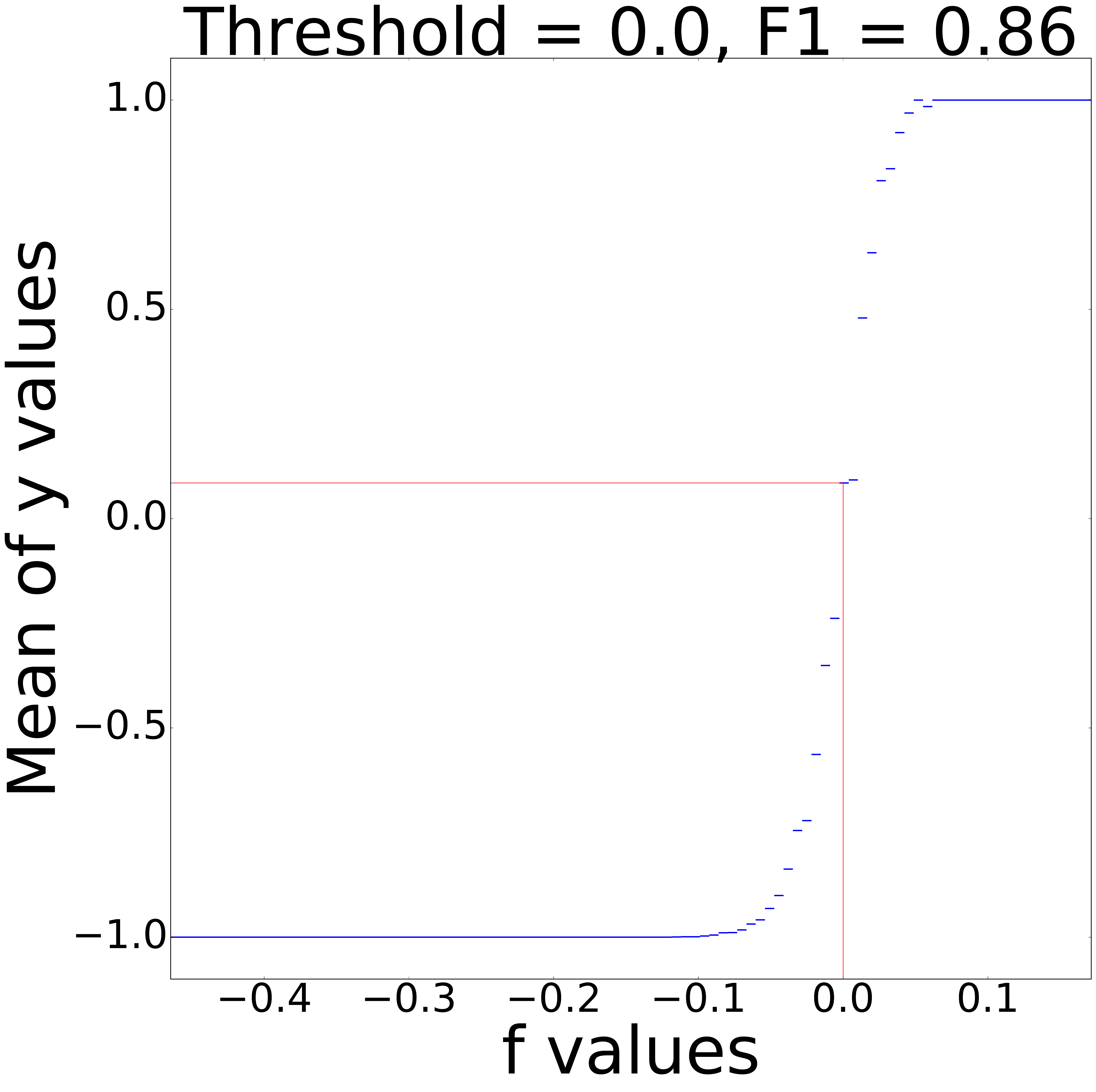}}
  \noindent\subfigure[Zeroth iteration]{\label{fig:makeup-1}\includegraphics[width=0.18\textwidth, height=0.09\textheight]{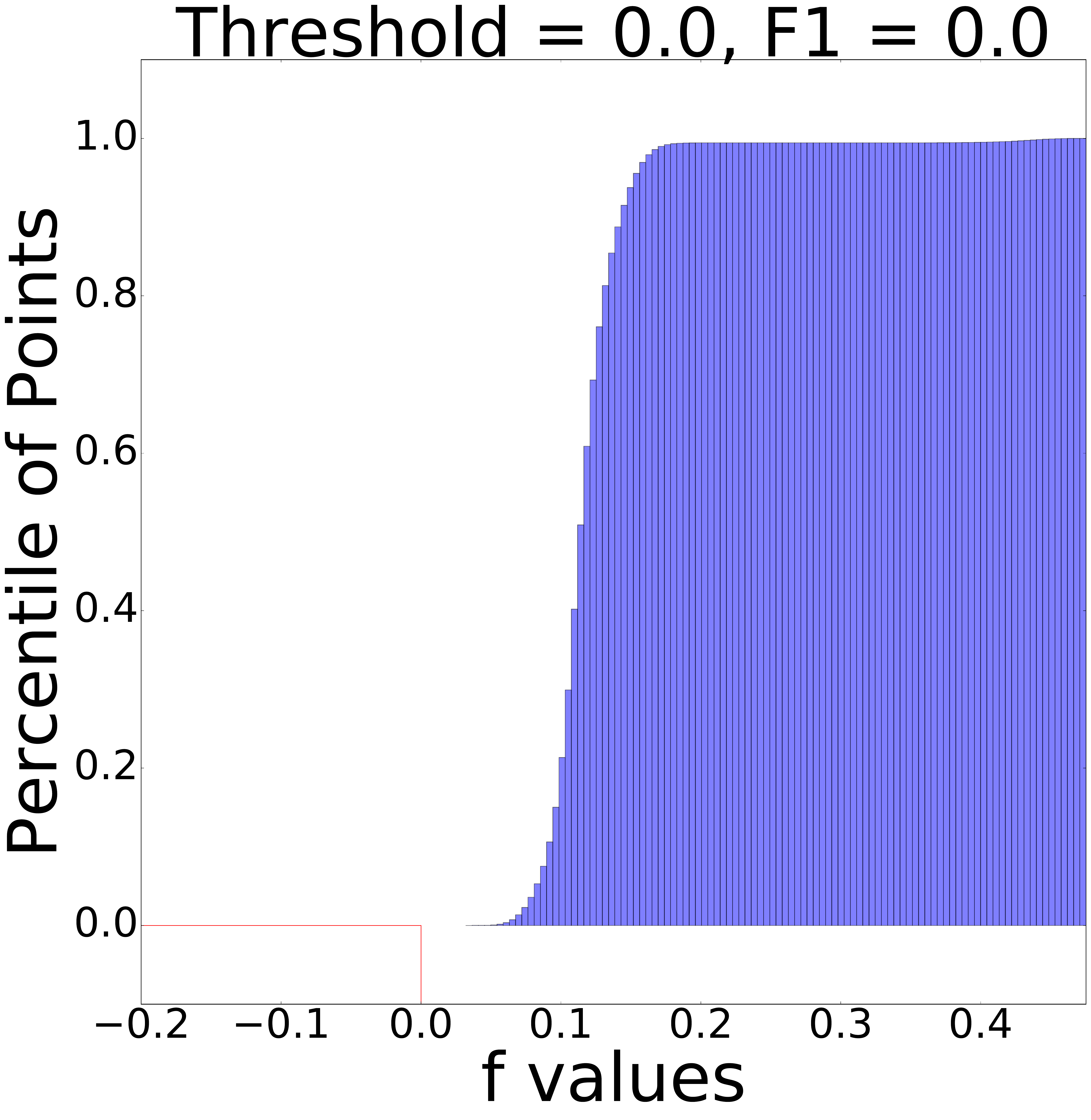}}
  \quad
  \noindent\subfigure[With adaptive threshold after 20 and 200 labeling rounds]{\label{fig:makeup-1}\includegraphics[width=0.18\textwidth, height=0.09\textheight]{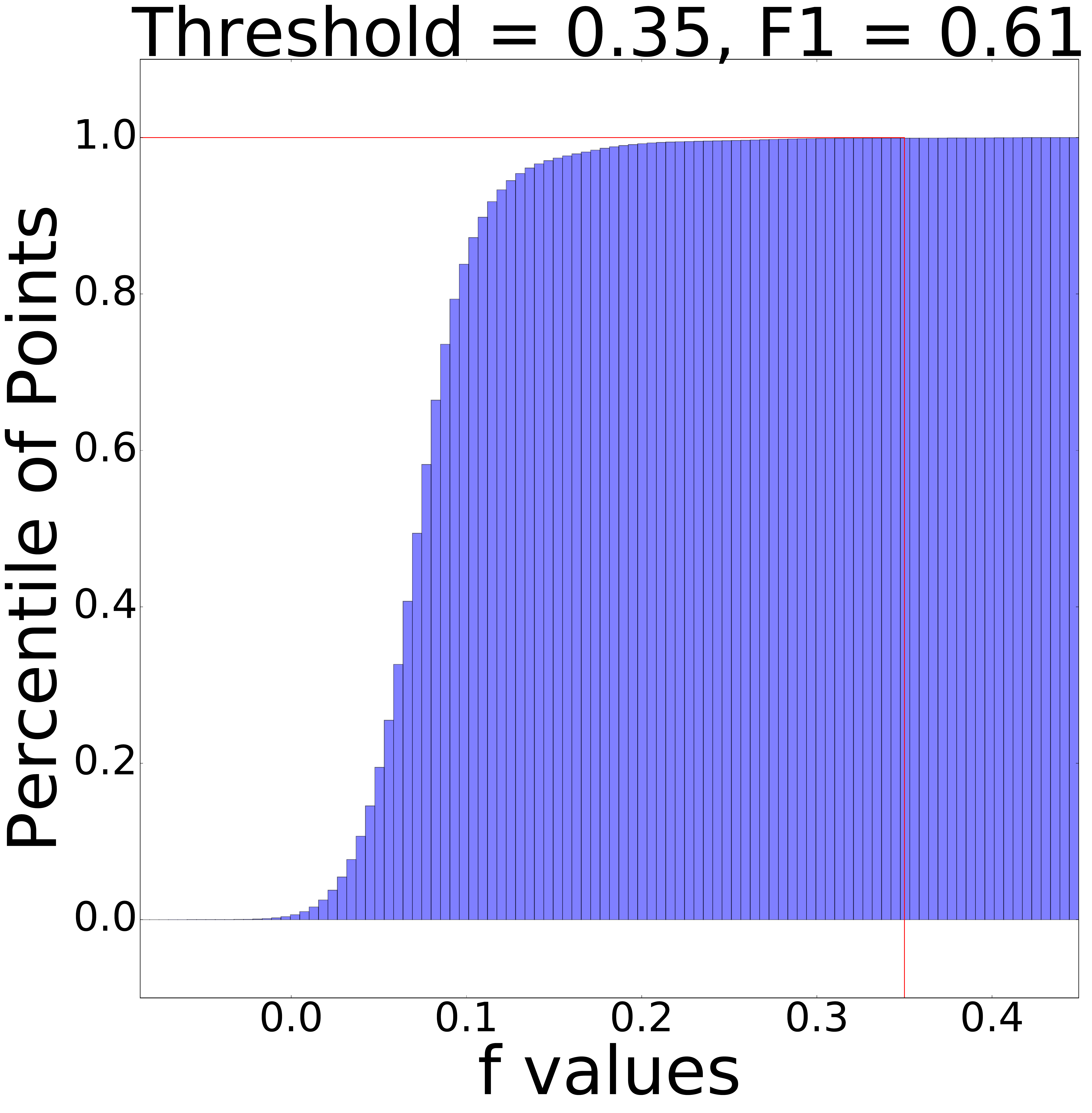}\includegraphics[width=0.18\textwidth, height=0.09\textheight]{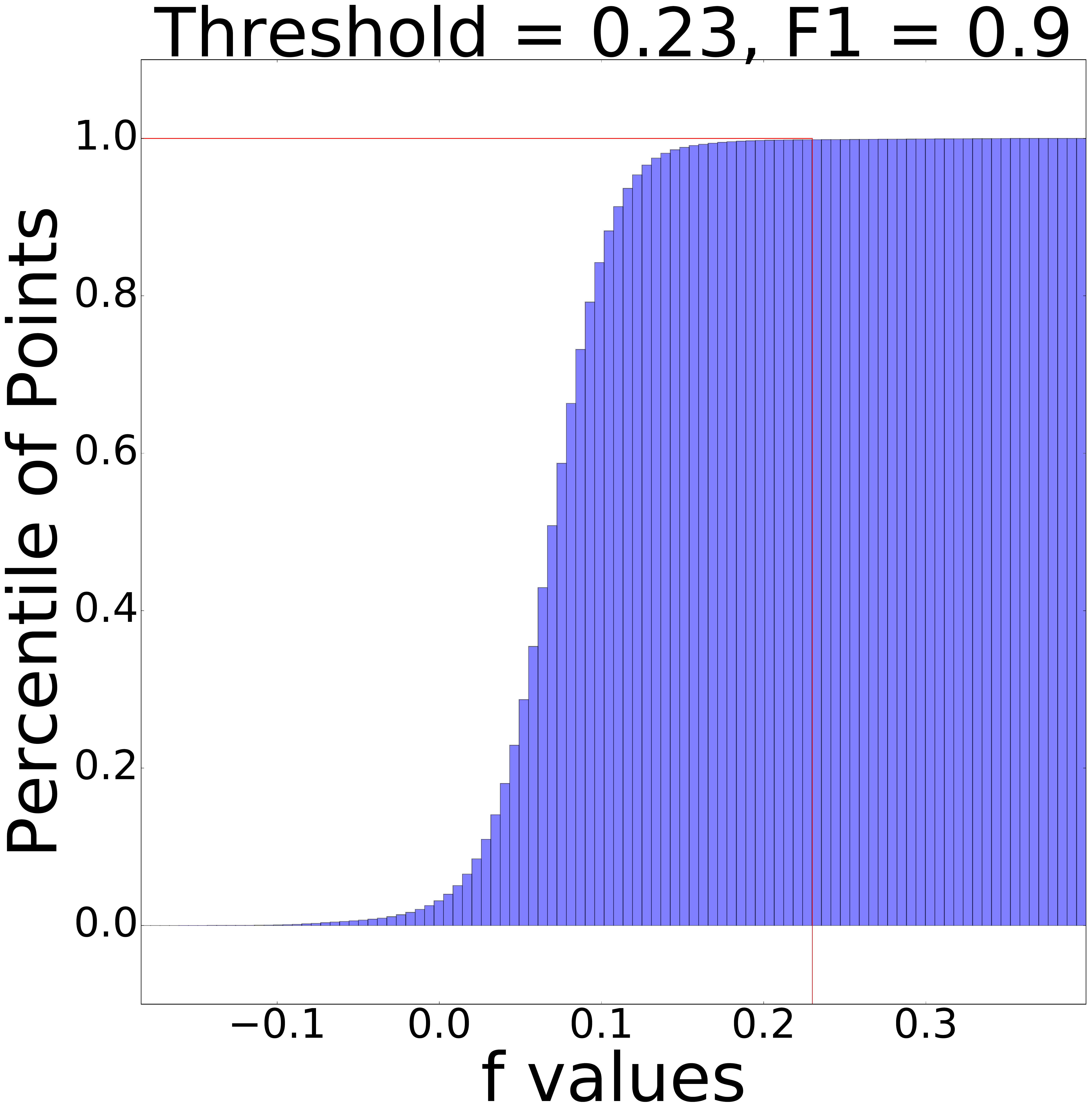}}
  \quad
  \noindent\subfigure[With adaptive threshold after 20 and 200 labeling rounds]{\label{fig:makeup-1}\includegraphics[width=0.18\textwidth, height=0.09\textheight]{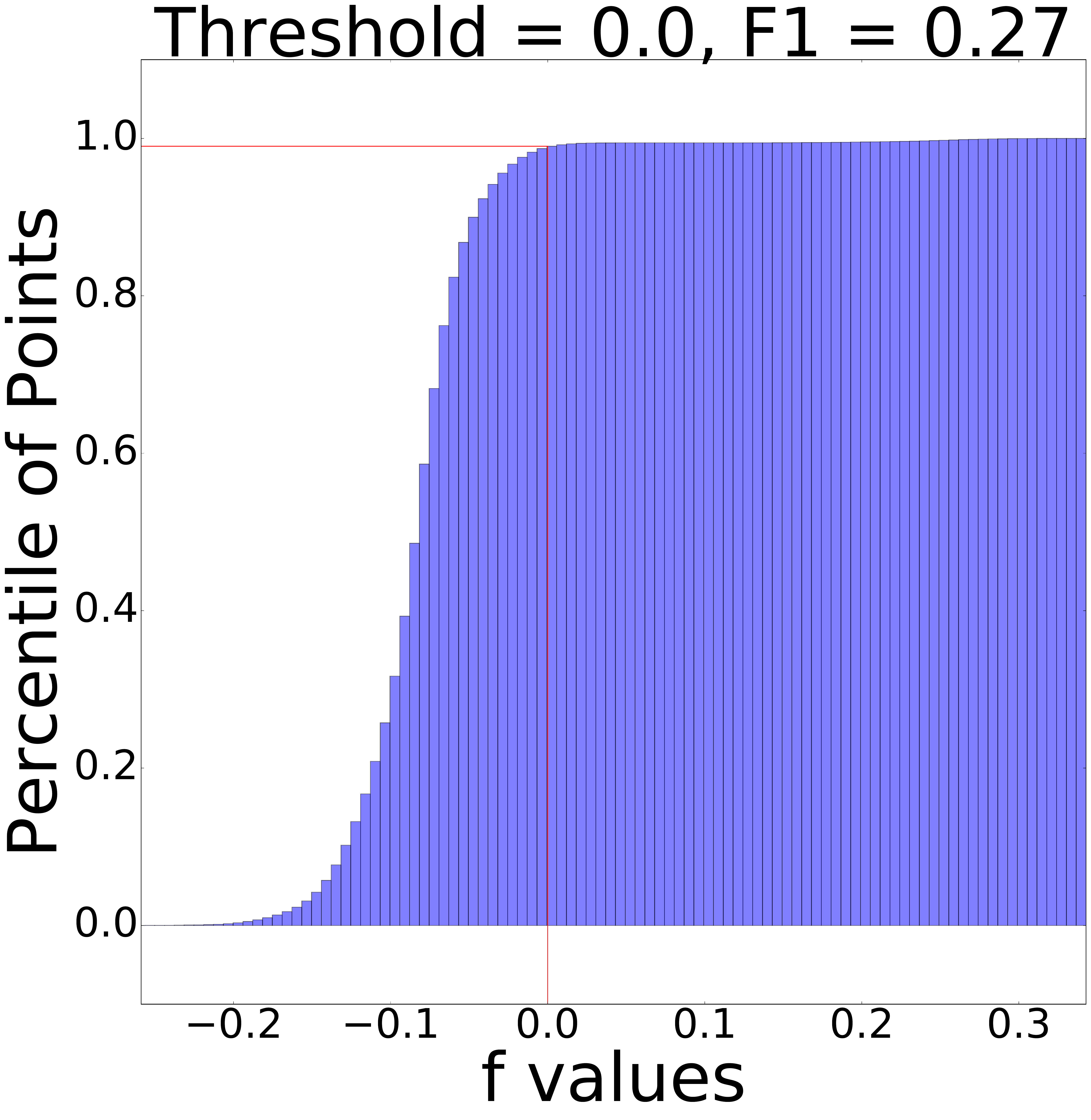}\includegraphics[width=0.18\textwidth, height=0.09\textheight]{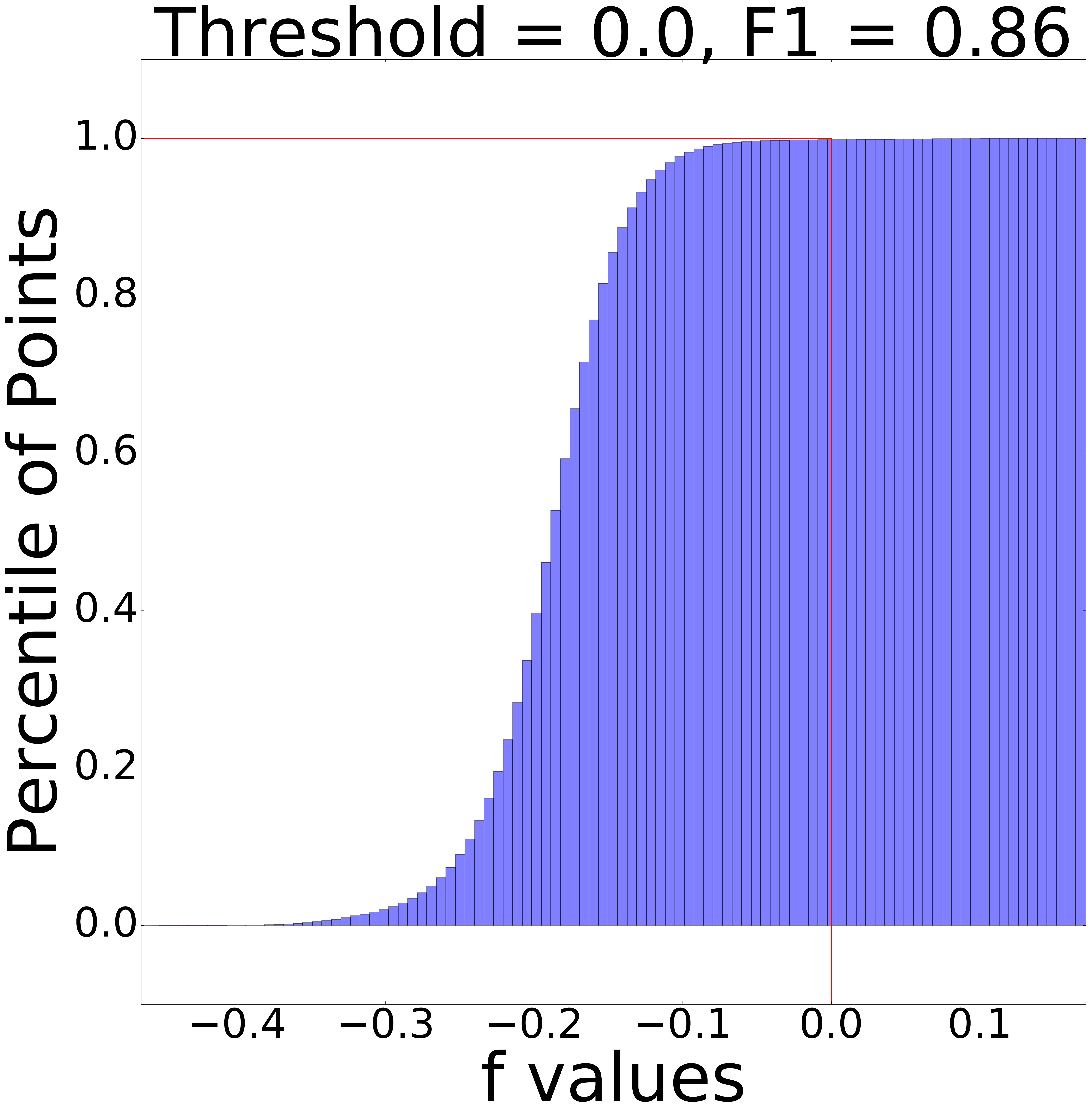}}
  \caption{Plot of mean label values(top) and cumulative distribution of the $f^*$ values(bottom), for the concept ``\textit{makeup accessories}'' with 10 labeled points after zero, 20, 200 rounds of active learning using adaptive(b, e) and constant(c, f) threshold}
  \label{fig:img-threshold}
  
\end{figure*}

\subsection{Adaptive threshold}
The choice of threshold is crucial for using the GSSL based methods for classification. The correct threshold should be indicative of the prior probability of the concept. We discuss two different techniques of choosing the threshold. The first is by simply setting it to a value of zero and the other is to adapt the threshold to better estimate the class prior. 

\subsubsection{Constant threshold}
The usual choice of threshold with the  $y_i \epsilon \{-1, 1\}$ is to set it to a constant such as zero. This is a reasonable choice when the true prior probability is nearly fifty percent. However, in case of concept detection from a large database, very few items actually belong to the concept user is looking for and thus the actual prior probability is extremely small. In such a case, choosing zero as the threshold is not a good choice. Figure \ref{fig:makeup} (right) shows an example of the cumulative distribution of $f^*$ values for an Imagenet concept, ``\textit{makeup accessories}''. The true prior for that concept is about 0.005, and the corresponding threshold should be about 0.2. If one uses the zero threshold ($f^*=0$), all images will be classified as positive (i.e., relevant) which is certainly not correct. In addition to the cummulative histogram, Figure \ref{fig:makeup} (left) shows the mean label value ($y \epsilon \{-1, 1\}$) of the corresponding points after binning the  $f^*$ values. We see that the points near the threshold ($f^*=0$) are mostly negative which shows that not only the threshold is wrong but also the new points chosen near this wrong decision boundary for the next round of active learning will not be very informative, since the current classifier has already assigned low $f^*$ values to these points. 

\begin{algorithm}
\caption{Algorithm for adaptive threshold}
\label{alg:alg-adaptive}
\begin{flushleft}
(Online steps)
\begin{algorithmic}
\State $i \Leftarrow 1$
\State $f^{*} \Leftarrow $ final $f^{*}$ for all features
\State $\Theta \Leftarrow 0$ 
\While{$i \neq $ Maximum number of active learning rounds}
\State $x \Leftarrow $ Point whose $f^{*}$ value is closest to $\Theta$
\State Query the label for $x$
\If{Predicted label of $x$ $\neq$ actual label of $x$}
  \If{Predicted label of $x$ $=$ -1}
  	\State $\Theta \Leftarrow \Theta - \frac{1}{\alpha  i}$.
  \Else
  	\State $\Theta \Leftarrow \Theta + \frac{1}{\alpha  i}$ 
  \EndIf
\EndIf
\State Add queried point to the set of labeled points
\State Obtain $f^{*}$ using the new set of labeled points
\State $i \Leftarrow i + 1$
\EndWhile
\end{algorithmic}
\end{flushleft}
\end{algorithm}

\subsubsection{Adaptive threshold}
As described earlier, an ideal threshold is one which can correctly determine the labels of all the points. But this is possible only if the classifier is highly accurate. Due to availability of only a few examples at the start of the search, the GSSL classifier is not very accurate initially. Examining figure \ref{fig:makeup} closely, we can see that a good choice of threshold is to choose the $f^*$ value corresponding to the bin with mean label value close to zero as this would be the bin which has equal number of positive and negative points in it. But, since we do not know the label values of all the points beforehand we need a method that can help us approximate such a threshold. We propose algorithm \ref{alg:alg-adaptive} to do this. The algorithm starts with an initial value of zero for the threshold and queries a point whose $f^*$ values is closest to it. If the predicted and actual values (obtained from the user) match, then no update is made to the threshold, otherwise the threshold is updated. If the predicted label of the queried point is negative when it should be positive, we lower the threshold to classify the point correctly and when the predicted label of the queried point is positive when it should be negative we increase the threshold. Doing this brings us closer to the actual decision boundary. Initially, we make larger adjustments to the threshold, but as the classifier becomes  accurate because of the active learning process, we need only small adjustments to the threshold as the $f^*$ values stabilize. 
In figure \ref{fig:img-threshold}, we contrast the adaptive threshold vs constant threshold. The figure 2(b) and 2(e) shows the effect of adaptive threshold, where we can see that the algorithm does well to move the threshold to the bin with zero mean label value. The increase in the F1 score suggests that the adaptive threshold is providing better classification results and also the points being queried during active learning are helping the classifier refine the concept boundary. This can be contrasted with the figure 2(c) and 2(f), that shows the effect of choosing points around zero, which actually hurts the learning of the classifier in the beginning. In figure \ref{fig:img-threshold}, starting with same labeled points and doing twenty rounds of active learning adaptive threshold provides much better F1 score (0.61) than obtained by using zero as the threshold (0.27). The classifier with zero threshold begins to catch up  with the adaptive threshold only after many iterations, which makes the proposed method superior to the constant threshold method in practice. An alternative to the incremental update in algorithm \ref{alg:alg-adaptive} is a bisection based search. However, using a bisection based method, there is a possibility that the correct value of the threshold might actually never be reached as the distribution of $f^*$ changes a lot during first few rounds of active learning. Due to these abrupt changes in $f^*$ values the figures \ref{fig:img-f1} and \ref{fig:svm-img-f1} are slightly rough. On the other hand, if the distribution of $f^*$ values remained fixed then the bisection method would be ideal to find the correct threshold in logarithmic time. 
\vspace{-0.1in}
\begin{algorithm}
\caption{Active Semi-Supervised Algorithm (Full)}
\label{alg:alg-active}
\begin{enumerate}
\item (Offline steps)
  \begin{algorithmic}[1]
	\item Compute \textit{visualEigenfunctions} using the offline steps of algorithm \ref{alg:alg-fergus}
	\item Compute \textit{semanticEigenfunctions} using the offline steps of algorithm \ref{alg:alg-fergus-modified}
	\end{algorithmic}
\item (Online steps)
	\begin{algorithmic}[1]
    \item Start with a few initial examples with labels
	\item Obtain $f^*_{visual}$ using the online steps of algorithm \ref{alg:alg-fergus} 
	\item Obtain $f^*_{semantic}$ using the online steps of algorithm \ref{alg:alg-fergus-modified} 
	\item Use algorithm \ref{alg:alg-adaptive} with 
    $f^*$ = $f^*_{visual}\lambda_{visual}$ + $f^*_{semantic}\lambda_{semantic}$ and $\lambda_{visual}$ + $\lambda_{semantic}$ = 1
    to adapt the threshold and query labels of informative points.
	\end{algorithmic}
\end{enumerate}
\end{algorithm}
\vspace{-0.1in}
\section{Experiments}
We use two different datasets for our experiments. The first is the AWA\cite{xian2017zero} dataset that contains about 37 thousand images of 50 animals. Each animal is annotated with 85 different attributes based on categories such as color, texture, shape, habitat etc. The second dataset is the down sampled version of the Imagenet\cite{deng2009imagenet} dataset \cite{chrabaszcz2017downsampled}. This dataset includes all images from the Imagenet, resized to 64x64. It has a total of 1.2 million images belonging to 1000 classes.

\subsection{Creating concepts from class labels}
We create concepts using original labels of the datasets by grouping the labels based on common attribute(s) which can either be visual, semantic or both. For the AWA dataset, we create 30 different concepts by grouping the animals based on attributes such as color, texture, shape and body parts. For example, a concept, ``\textit{big animals with black stripes}'' includes animals like ``zebras'' and ``tigers''. For Imagenet, we rely on the WordNet hierarchy to create concepts. We merge the leaf level nodes into their common ancestor based on the hierarchy and then use this ancestor node as the concept. For example, the labels ``daisy'' and ``yellow lady's slipper orchid'' have a common ancestor ``\textit{flower}'' and thus ``\textit{flower}'' is a concept. We create 30 different concepts in this way. Since a typical retrieval task in a large database,  has only a handful of relevant results, we ensure that all the concepts which we test on have at most $<5\%$ (< 1800 out of 37,000) positive points for AWA and $<1\%$ (< 12,000 out of 1,200,000) for Imagenet.

\subsection{Feature extraction}
We use pre-trained deep learning networks like Resnet\cite{he2016deep} and Xception\cite{chollet2016xception} for visual features. To extract the features for the Imagenet dataset we pass all the images through a pre-trained Xception network available in Keras \cite{chollet2015keras}. The input to the Xception network is an image of size 299x299. We resize our images to this size and obtain the output of 2048 dimensions. We use these features directly for the Imagenet dataset. For AWA dataset, we extract the features from the Resnet network using a similar procedure. Then we use these features in a multi-task learning setting to train a neural network that can detect different attributes. We use features from the shared layer as visual features for AWA dataset. We use the WordNet hierarchy for obtaining semantic features. Since the Imagenet dataset is built using the WordNet hierarchy, each class label has a synset associated with it. Likewise, for AWA, we map each animal to its nearest synset in the WordNet. We use these synsets to compute semantic features. Specifically, we use the Lin's similarity\cite{budanitsky2001semantic} measure to construct the affinity matrix as described in section 4.2. 

\subsection{Train and test splits}
For the Imagenet dataset, we randomly select 500 images belonging to each class label (leaf nodes) and use those to form the test set. For AWA, we randomly choose 100 images from each of each class label for our test set. Thus, test set size for Imagenet is 500,000 images and is 5000 images for AWA. Performance evaluation (F1 scores) is based on the images in the test set. Remainder of the images act as a big pool of unlabeled data. The user labels a few images from this pool and starts the search. The system uses the remaining images from this pool of unlabeled data for active learning.

\subsection{Computing the eigenfunctions}
Here we describe the implementation details for computing the eigenfunctions for visual and semantic features. For eigenfunctions of the visual features, we first use PCA to reduce the dimensionality of the features to 64 from 2048 dimensions in Imagenet and 512 dimensions in AWA. Then we use offline steps of algorithm \ref{alg:alg-fergus} to compute the eigenfunctions by discretizing the density into 500 bins and computing the eigenfunctions for a single axis using equation 2. We repeat this procedure 64 times and get eigenfunctions for each axis in the input. Then we sort the eigenvalues obtained over all the axes and discard the eigenfunctions corresponding to the nearly zero eigenvalues (< $10^{-10}$), as these eigenfunctions correspond to the constant solution for equation 2. After discarding the eigenfunctions corresponding to extremely small eigenvalues, we keep 256 eigenfunctions corresponding to the smallest eigenvalues (roughly 4 per dimension) for both Imagenet and AWA. Then we interpolate all the data using these 256 eigenfunctions. To compute eigenfunctions of the semantic features, we follow the offline steps of algorithm \ref{alg:alg-fergus-modified}. We use the affinity matrix created using the WordNet Hierarchy and use it to solve Equation 2. We keep 100 eigenfunctions for Imagenet and 10 eigenfunctions for AWA after discarding the eigenfunctions corresponding to extremely small values. Using these eigenfunctions we interpolated the rest of the data. It takes $<1$ minute for visual eigenfunctions and $<10$ seconds for semantic eigenfunctions on a single core CPU machine. 

\begin{figure}[t]
\centering
\noindent\subfigure[AnimalWithAttributes]{\label{fig:awa-f1}\includegraphics[width=0.22\textwidth, height=0.13\textheight]{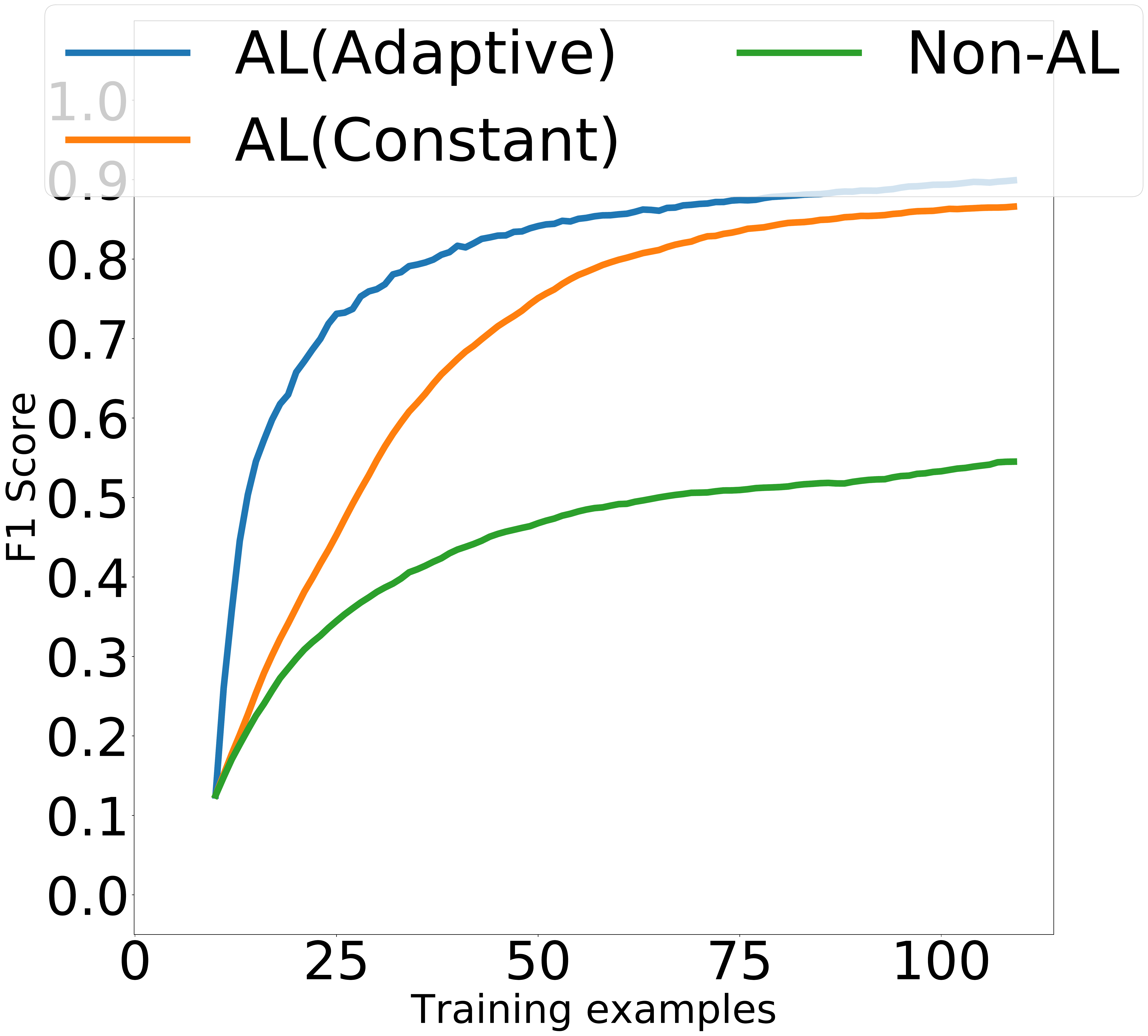}}
    \quad
\noindent\subfigure[Imagenet]{\label{fig:img-f1}\includegraphics[width=0.22\textwidth, height=0.13\textheight]{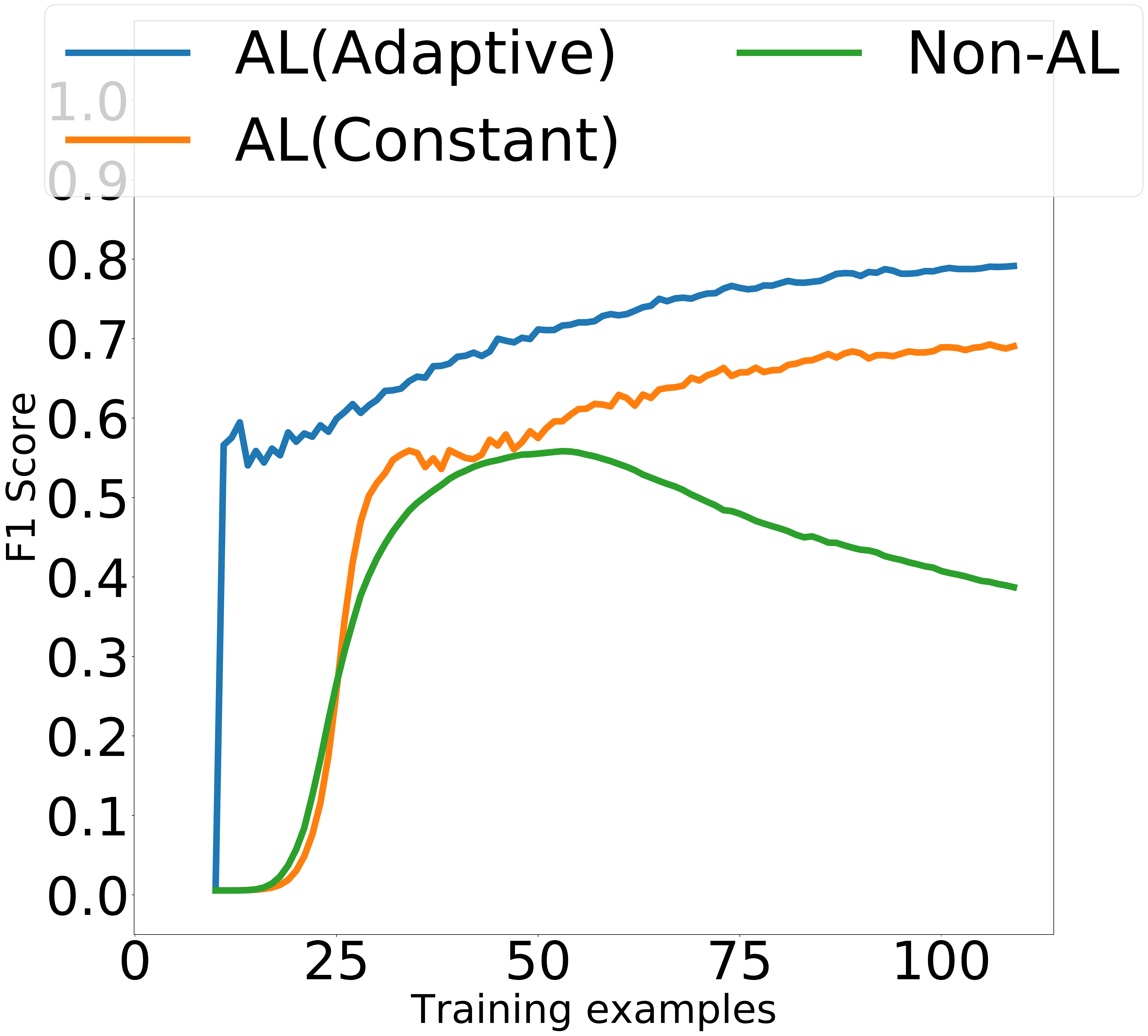}}

\caption{F1 Scores for 30 concepts defined on the AWA (left) and Imagenet (right) dataset averaged over 20 times. The graph contrasts the method with and without using active learning(Non-AL). It also compares the active learning methods with and without using adaptive threshold.}
\label{fig:scores}
\end{figure}

\subsection{Computing $f^*$}
In order to compute the initial label functions, we randomly label 10 images, 9 that belong to the concept and 1 that does not. This is equivalent to the user starting the search, where they would provide some labels for images which they are interested in seeing. Using these labeled images we compute the label function for visual and semantic features and by solve equation 1 as per the online steps of algorithm \ref{alg:alg-fergus} for visual features and algorithm \ref{alg:alg-fergus-modified} for semantic features. Once we have the label functions ($f^*_{visual}$ and $f_{semantic}^*$) for both the features, we combine the individual predictions as described in section 4.3. This process takes $<5$ seconds on a single core CPU machine. We use $\lambda = 0.5$ while combining the predictions. This value worked well for the concepts we tested on and is equivalent to giving equal weights to the information from both visual and semantic features. We evaluate the performance of the method by measuring the F1 scores for all the concepts averaged over 20 runs.

\subsection{Using Active Learning}
\begin{figure}[t]
\centering
\noindent\subfigure[AnimalWithAttributes]{\label{fig:svm-awa-f1}\includegraphics[width=0.22\textwidth, height=0.13\textheight]{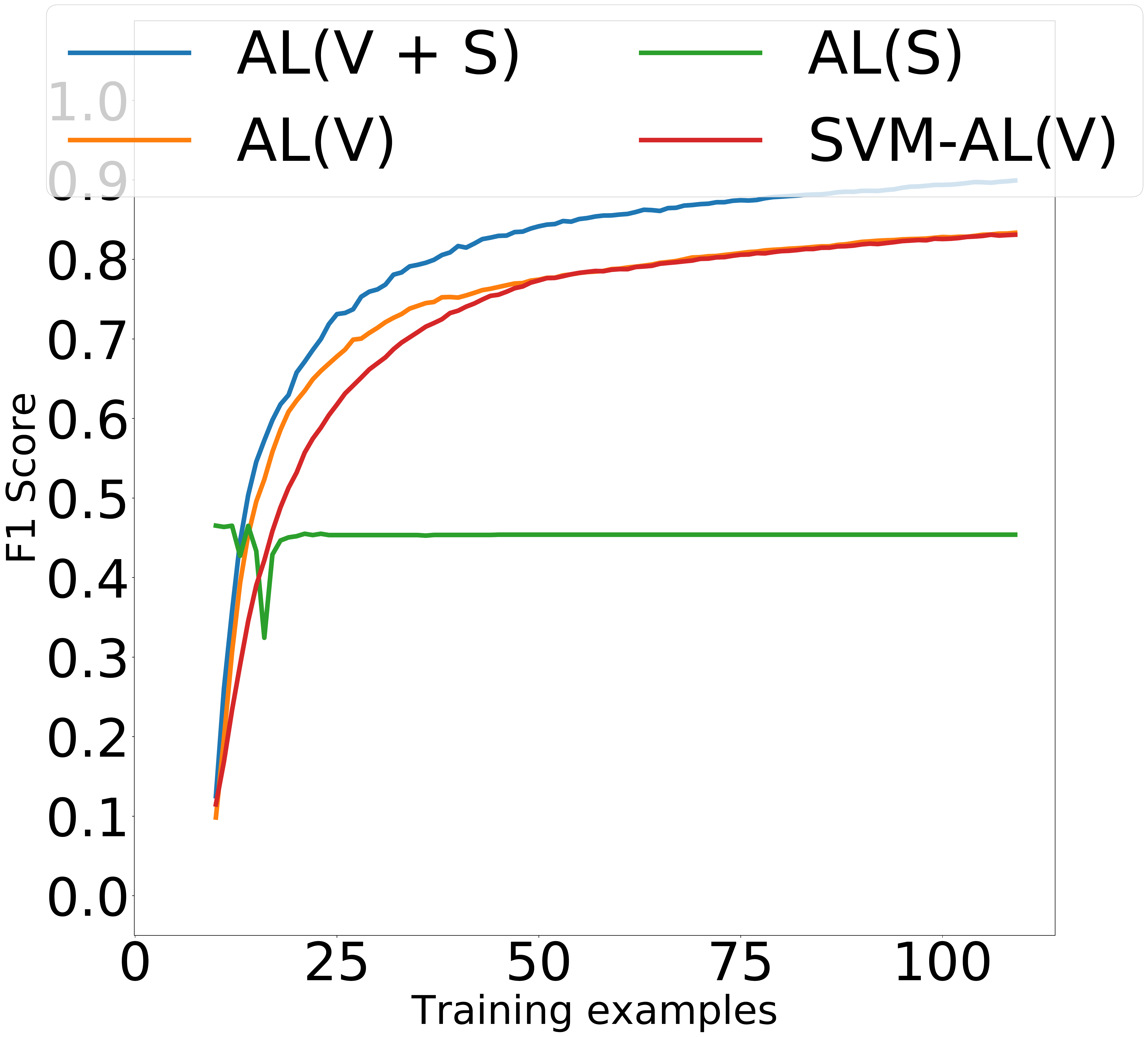}}
    \quad
\noindent\subfigure[Imagenet]{\label{fig:svm-img-f1}\includegraphics[width=0.22\textwidth, height=0.13\textheight]{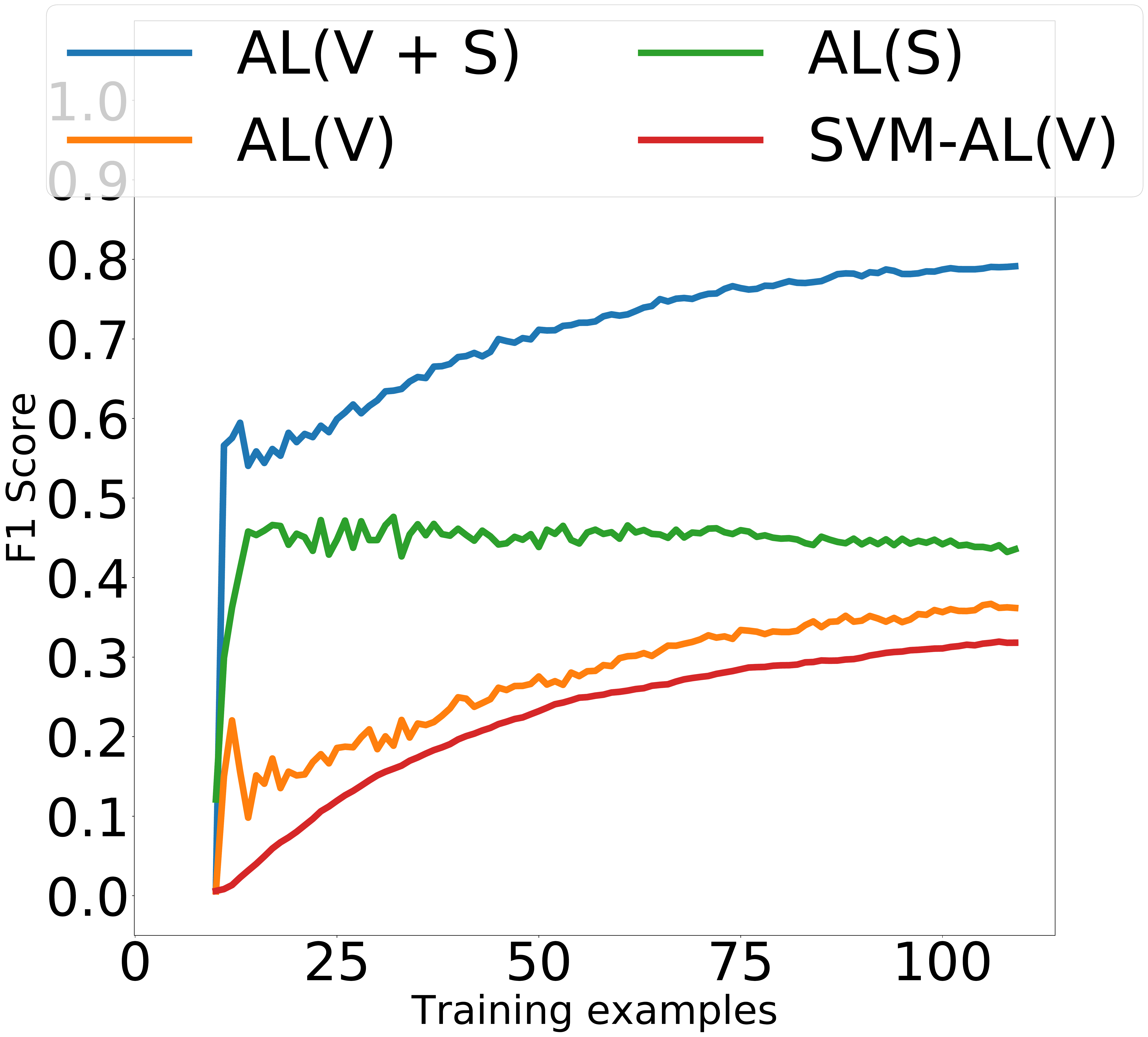}}

\caption{Comparison of our method against SVM with active learning. We contrast the performance of SVM with visual(V) features against our method  using only visual features, only semantic(S) features and using both visual and semantic features.}
\label{fig:svm_scores}
\end{figure}

As described in section 5, we use active learning to query the labels of the most informative points. These are the points which are near the decision boundary. We follow algorithm \ref{alg:alg-adaptive} to find the threshold which acts as the decision boundary and then select points closest to it. For our experiments, the threshold is initialized to zero at the start and $\alpha = 2$ (based on cross validation). We start our experiment by labeling 10 images and from then on label only a single image in each round of active learning. The classifier is retrained with this newly added point. The results in figure \ref{fig:scores} show the performance of three techniques of adding points to the labeled set. The first method is random sampling where the system queries a random point to be labeled. We see that adding a random point does not help the learning of the classifier after a certain stage, in fact, it actually hurts performance. This is because random sampling has a higher probability of adding a negative point to the labeled set, as there are a large number of irrelevant images and only a few relevant images. Due to the addition of many negative points, the influence of the positive points in the graph decreases eventually, rendering all predictions as negative. Since Imagenet has many more negative points, this effect is prominent in figure \ref{fig:img-f1}. The other two methods are based on active learning, with and without the use of adaptive threshold. We see both the active learning methods perform considerably better than the random selection. The performance of active learning methods using the adaptive threshold is significantly better than the one which samples points around a constant threshold. The steep increase in the F1 scores with using adaptive threshold for both AWA and Imagenet datasets indicates that the method with adaptive threshold learns the concept with only a few rounds of interaction with the user. We see in figure \ref{fig:img-f1}, the F1 score for adaptive threshold goes from nearly zero to a high value. This is because the threshold is initially set to zero and after the first round of active learning when adaptive threshold takes over it boosts the performance. Although, both methods of active learning give similar performance in the long run, no user would like to do more than a few rounds of active learning. Thus, the initial region of the graph is considered the most important. In figure \ref{fig:svm_scores} we present comparison of our method against SVM with active learning. Since it is difficult to incorporate graph-type information directly in SVM we evaluate it only using visual features. We query the points nearest to the decision boundary for active learning. Our experiments show that our approach with only visual features performs better than SVM on both datasets and is also significantly faster. Combining both visual and semantic information provides significant gains to our approach over SVM. Additional experimental results are available at http://bit.ly/2EfjXf2.   

\begin{figure*}
  \begin{center}
  \subfigure[\textbf{Concept: ``\textit{Person playing a wind instrument}''} Images must show people playing different wind instruments ]{\label{fig:img-img1}\includegraphics[width=\textwidth, height=0.14\textheight]{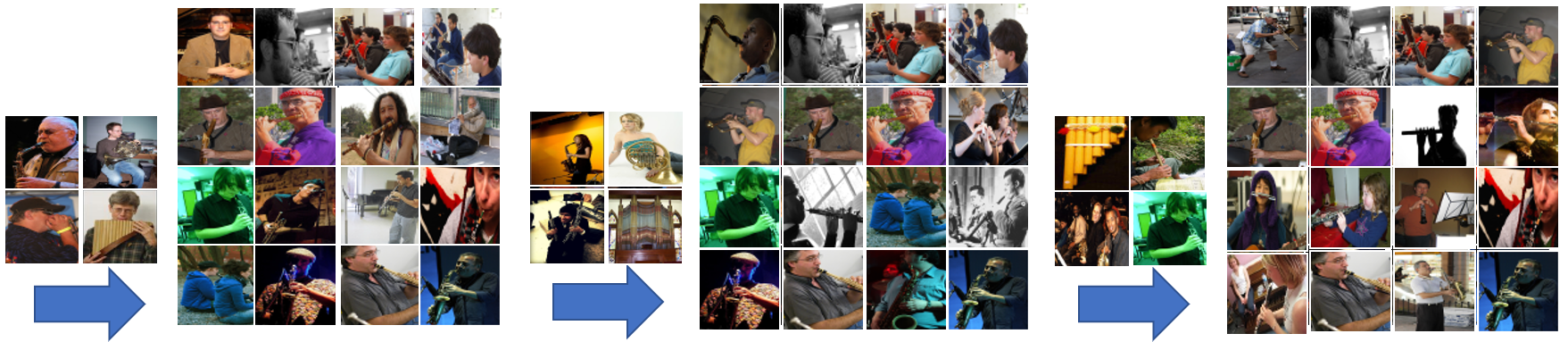}}
   \subfigure[\textbf{Concept: ``\textit{Group of animals}''} Images retrieved must have more than one animal in it. ]{\label{fig:img-awa1}\includegraphics[width=\textwidth, height=0.14\textheight]{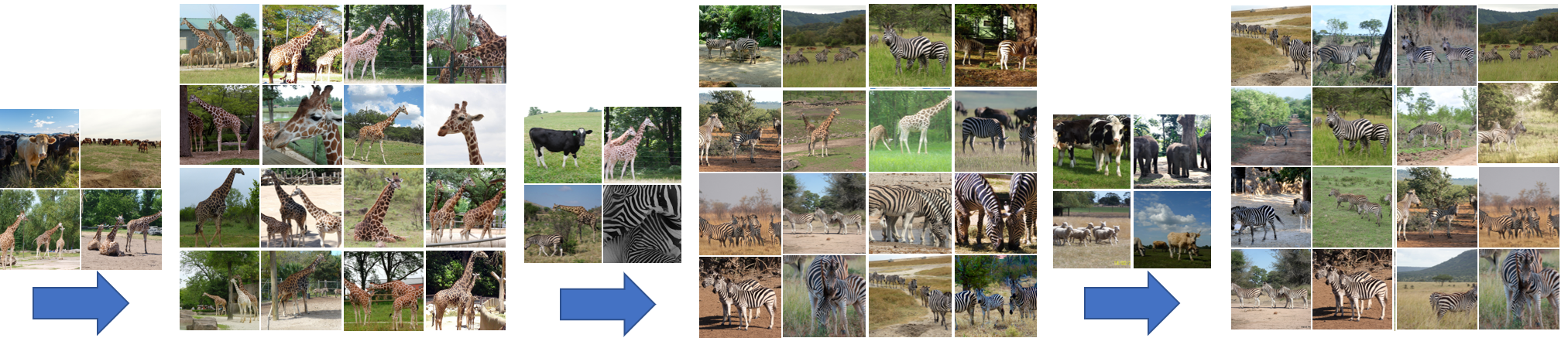}} 
    \subfigure[\textbf{Concept: ``\textit{Keyboard instruments}''} Images of instruments such as accordion, piano, upright are considered relevant.]{\label{fig:img-img2}\includegraphics[width=\textwidth, height=0.13\textheight]{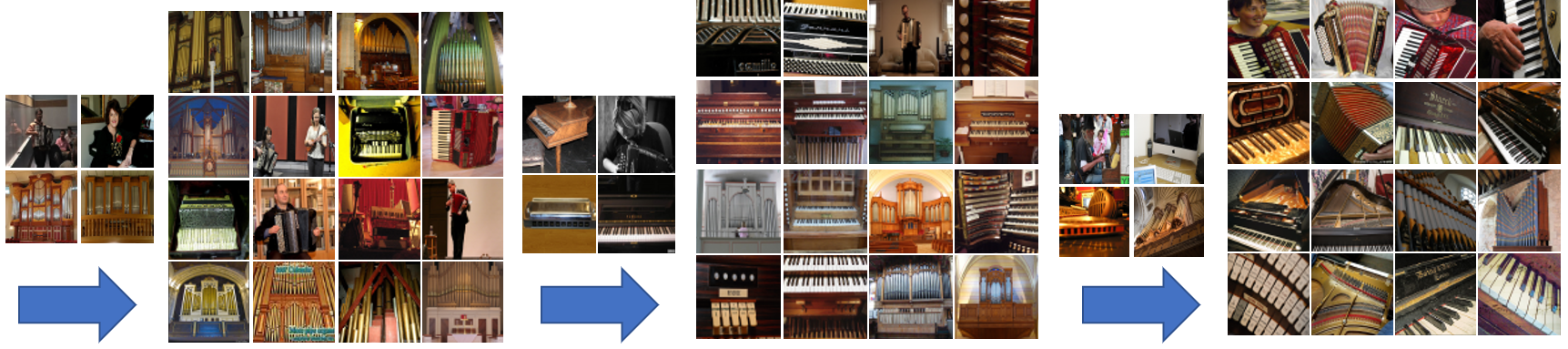}} \\
    \subfigure[\textbf{Concept: ``\textit{Brown bulbous animals with horns}''} Images of animals such as buffaloes, oxes etc. are considered relevant]{\label{fig:img-awa2}\includegraphics[width=\textwidth, height=0.13\textheight]{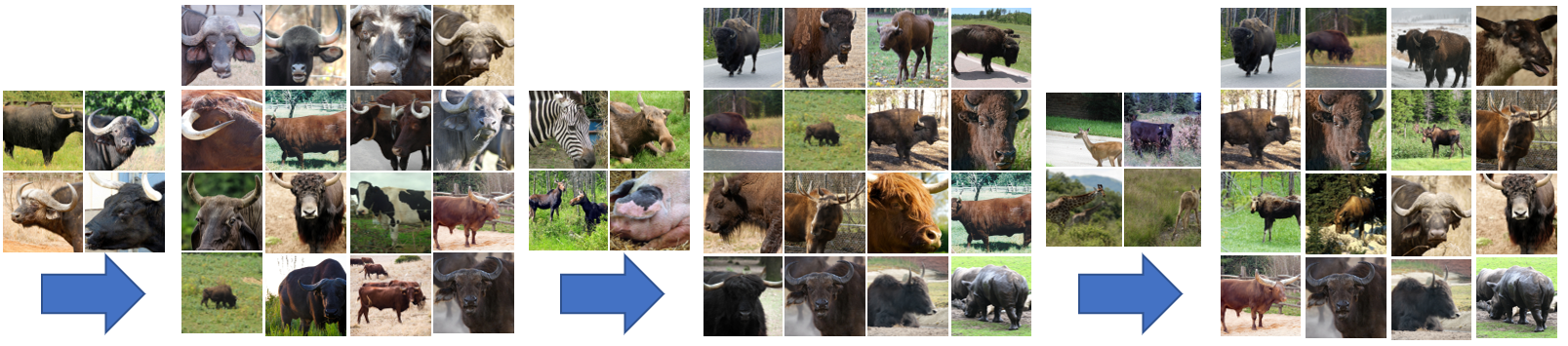}} \\ 
    \subfigure[\textbf{Concept: ``\textit{Makeup accessories}''} Images of makeup accessories such as face powder and lipstick are considered relevant.]{\label{fig:img-img3}\includegraphics[width=\textwidth, height=0.13\textheight]{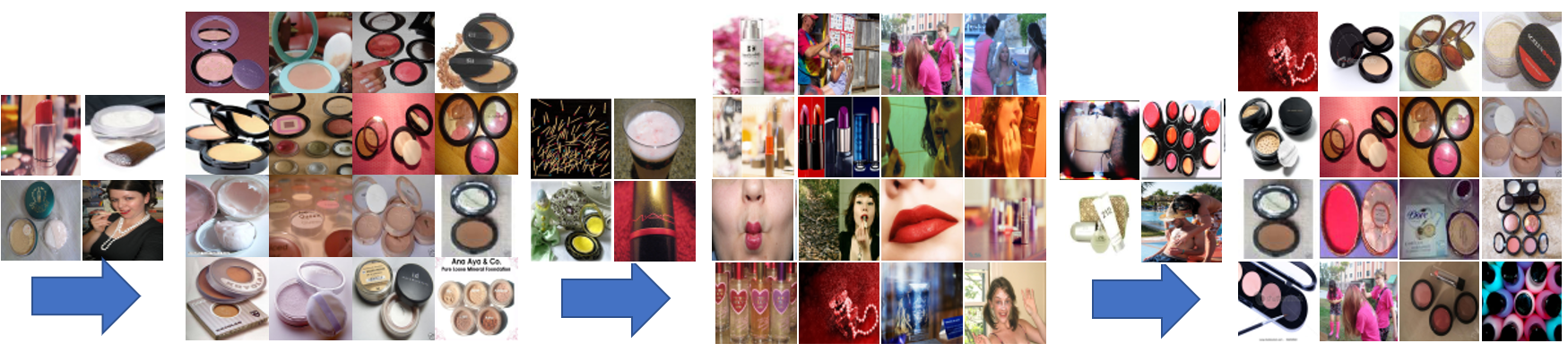}}
  \subfigure[\textbf{Concept: ``\textit{Furry animals with black stripes}''} Images of animals such as racoons, zebras, tigers etc. are considered relevant.]{\label{fig:img-awa3}\includegraphics[width=\textwidth, height=0.13\textheight]{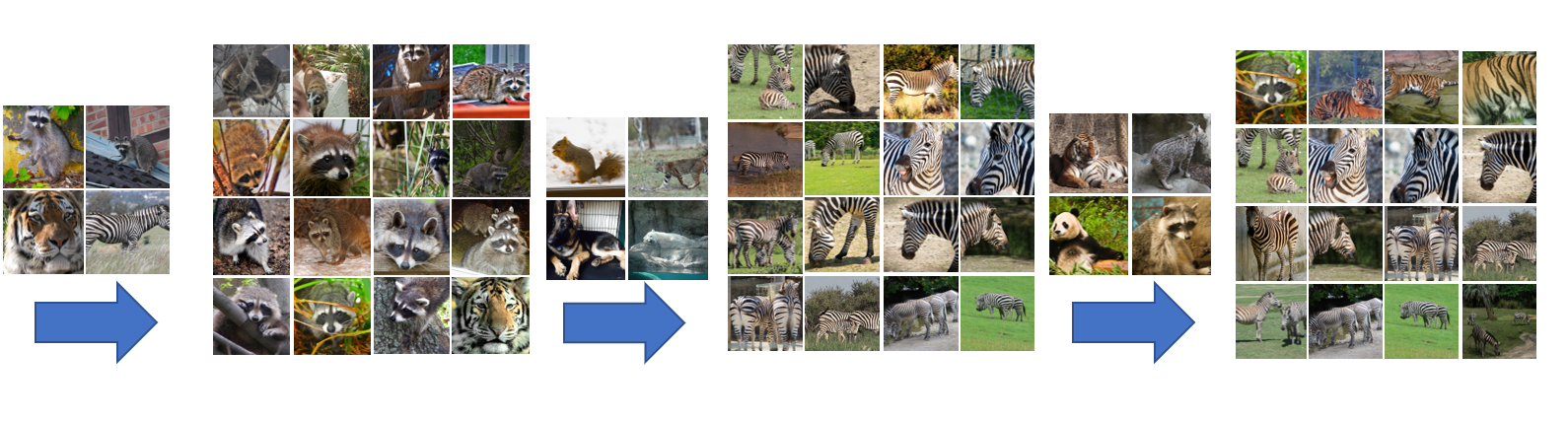}}
  \end{center}
  
  \caption{Retrieving images for various concepts defined on the Imagenet and AWA dataset. In images (a)-(f), the images above the first arrow are the images provided by the user to start the search, images over the second and the third arrow are images queried by the system for labels during active learning. Remaining images show top-16 results retrieved by the system after 0, 4 and 8 rounds of active learning.}
  \label{fig:imagenet}
\end{figure*}

\subsection{Use case examples}
The results for retrieval of various concepts using our method are present in figure \ref{fig:imagenet}. The user starts by providing some examples images to the system. Using visual and semantic information from these images $f^*$ is computed. Images with highest $f^*$ are shown to the user along with an image near the decision boundary for labeling. Using the newly labeled image $f^*$ is recomputed. This process continues till the user is satisfied with the results. Figures \ref{fig:img-img1}, \ref{fig:img-img2} and \ref{fig:img-img3} show retrieval results for concepts ``\textit{a person playing a wind instrument}'', ``\textit{keyboard instruments}'' and ``\textit{makeup accessories}'' defined on Imagenet. For the concept ``\textit{makeup accessories}'' the system must find images of face powder, lipstick etc. and for the concept ``\textit{keyboard instruments}'' it must retrieve images of organ, grand piano, accordion etc. The concept ``\textit{person playing a wind instrument}'' is a difficult one as it requires the system to retrieve images of all wind instruments (semantic) with a condition that images must have a person playing that instrument (visual). Good results for this concept indicate that the system is using both semantic and visual information.
Figures \ref{fig:img-awa1}, \ref{fig:img-awa2} and \ref{fig:img-awa3} shows ranking for concepts defined on AWA. For the concept ``\textit{furry animals with black stripes}'', images of animals such as zebras, tigers, raccoons, skunks are relevant and for the concept ``\textit{brown bulbous animals with horns}'' images of moose, oxes etc. are relevant. The concept ``\textit{group of animals}'' requires images retrieved to have more than one animal. Our results clearly show that the system is able to retrieve images of several different animals (semantic) present in a group (visual).   
For all concepts, it can be clearly seen that after just 8 rounds of labeling i.e. having about 12 labeled images the top ranked results start to precisely reflect the concept. It takes less than three minutes for the system to reach this state (including the time for computing the ranking 9 times and the time taken by the user to label 12 images) on a single core CPU machine with 64 GB memory. This attests that our method is suitable for retrieval from large databases.

\section{Conclusion and Discussion}
In this work, we propose a fast and scalable method that combines active learning with efficient GSSL to learn a user's query concept with minimal user interaction. We presented how to use features of different modalities in GSSL framework, by constructing separate graphs and then obtaining the final predictions as a convex combination of individual predictions. We showed that points selected based on the adaptive threshold are the most informative ones and help the classifier learn the query concept quickly. Good results on Imagenet and AWA datasets are a concrete proof of effectiveness of our method for a large scale system. In the future, we plan to explore applicability of different active learning methods to large scale problems.

\bibliographystyle{ACM-Reference-Format}
\bibliography{sample-bibliography} 

\appendix
\section{Appendix}
\subsection{Names of concepts defined on Imagenet}
The following are the different concepts we defined on the Imagenet dataset for evaluating our method:\\
\textit{wheel, makeup, poodle, elephant, shore, hosiery, setter, fox, wolf, bear, free-reed instrument, source of illumination, flower, heron, soft-finned fish, coraciiform bird, bridge, domestic cat, crocodilian reptile, bowl, guitar, piece of cloth, sled dog, thrush, sailboat, seabird, stork, citrus, frozen dessert, piano}. 

\subsection{Names of concepts defined on AWA}
The following attributes from the list of 85 attributes were combined to create different concepts on AWA:\\
\textit{blue hairless strain teeth, blue tough skin bulbous, blue tough skin flippers, brown hairless hooves, orange spots quadrupedal, white spots flippers, yellow spots claws, white spots small, brown tough skin flippers, black tough skin hands, blue hairless tail, brown hairless flippers, brown spots claws, yellow furry big, white spots claws, yellow spots meat teeth, black hairless small, blue hairless big, white stripes small, black hairless strain teeth, black stripes meat teeth, brown tough skin hands, white stripes paws, orange spots lean, brown tough skin claws, white hairless hooves, orange furry chew teeth, black tough skin strain teeth, brown spots small, black spots lean}

\subsection{Mean Average Precision Curves}
A precision/recall curve plots precision and recall for every all possible thresholds. The curve decreases regularly between the points of (higest precision, lowest recall) and (lowest precision, highest recall). A slow decreasing curve is considered ideal. Mean average precision is a value that summarizes the precision/recall curve. This value is equivalent to the area under the precision recall curve and is independent of the threshold. Here we report the change in average precision scores after adding a single point queried by active learning. Figure \ref{fig:ap-scores} contrasts the performance of our active learning method with adaptive threshold against active learning with constant threshold and not using active learning at all for AWA and Imagenet datasets. The comparison suggests that our method of using active learning with adaptive threshold is better than the other methods. Figure \ref{fig:ap-scores} compare the performance of our method against only visual, only semantic and with the combination of visual and semantic features against SVM with only visual features. Our method performs well here too.
\begin{figure}[!htb]
\centering
\noindent\subfigure[AnimalWithAttributes]{\label{fig:awa-ap}\includegraphics[width=0.22\textwidth, height=0.15\textheight]{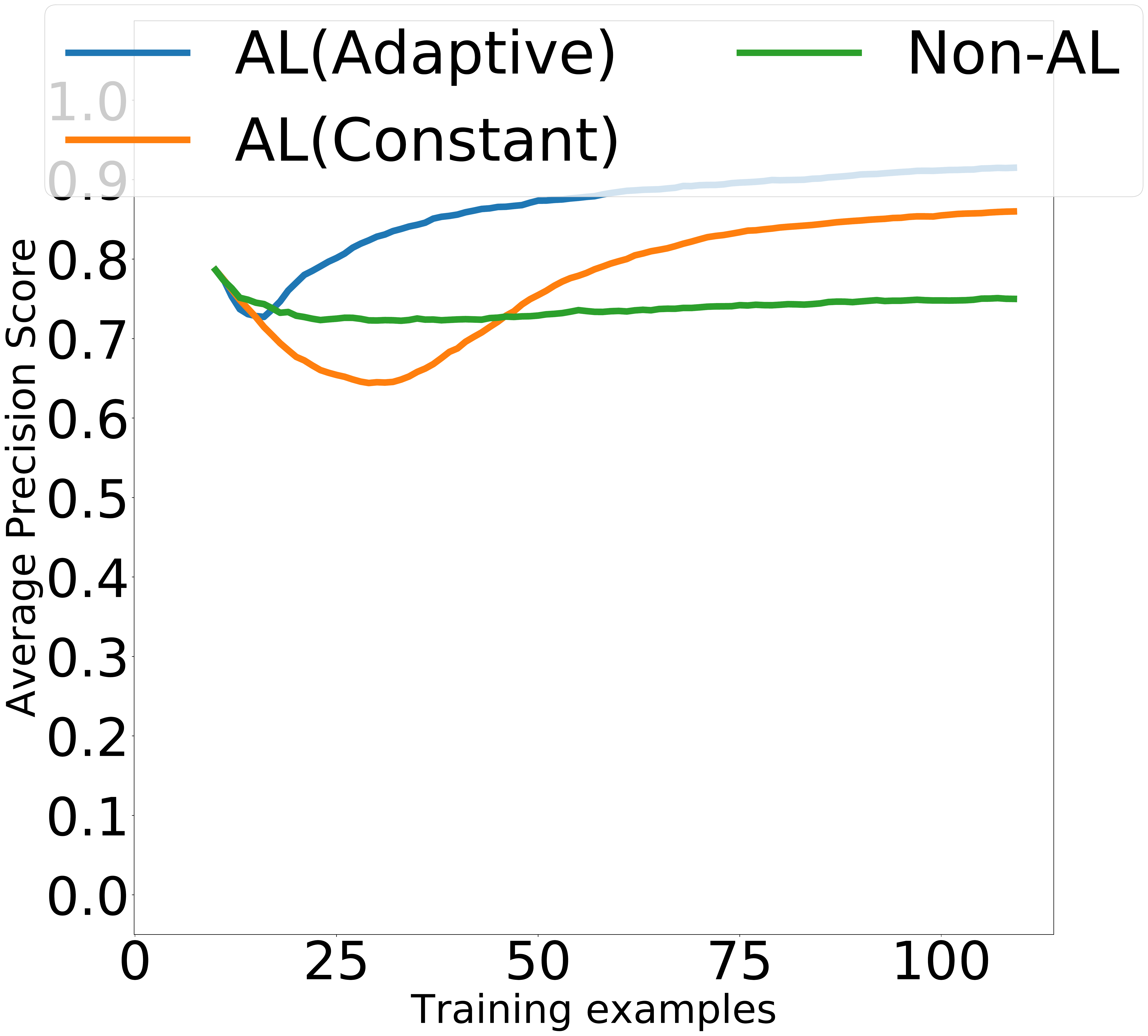}}
    \quad
\noindent\subfigure[Imagenet]{\label{fig:img-a}\includegraphics[width=0.22\textwidth, height=0.15\textheight]{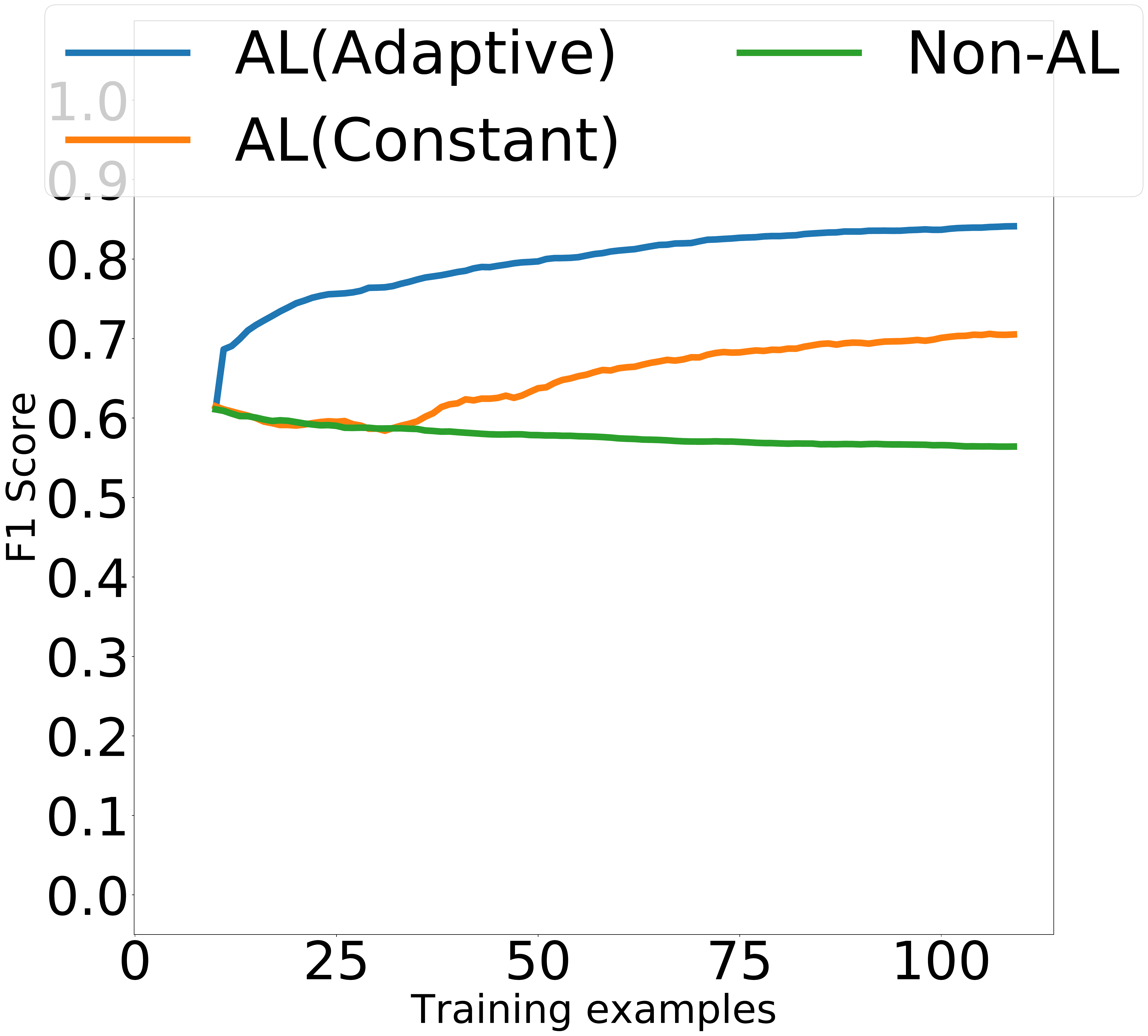}}

\caption{Average Precision Scores for 30 concepts defined on the AWA and Imagenet datasets. The graph contrasts the method with and without using active learning. It also compares the active learning methods with and without using adaptive threshold.}
\label{fig:ap-scores}
\end{figure}

\begin{figure}[!htb]
\centering
\noindent\subfigure[AnimalWithAttributes]{\label{fig:awa-svm-ap}\includegraphics[width=0.22\textwidth, height=0.15\textheight]{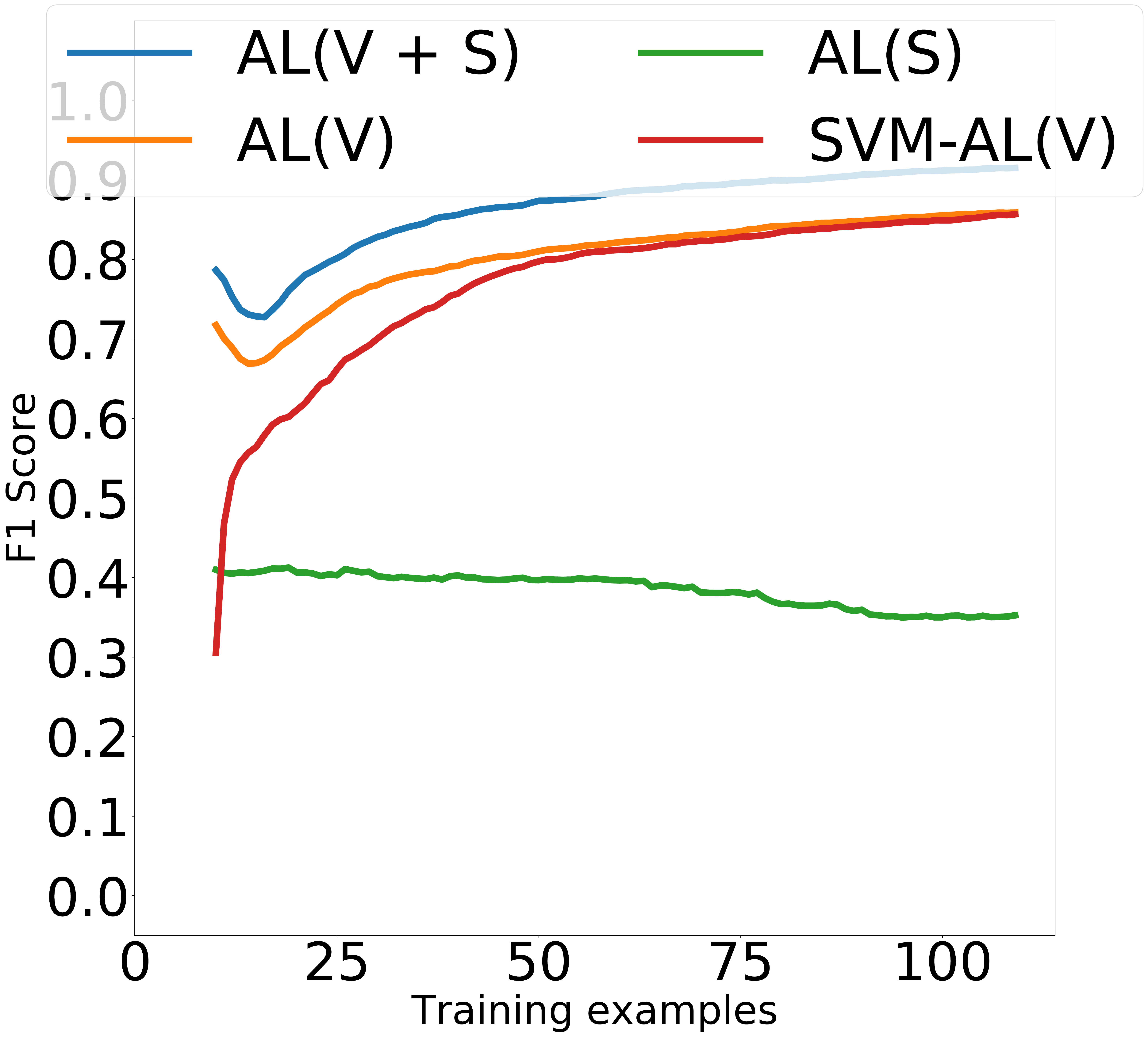}}
    \quad
\noindent\subfigure[Imagenet]{\label{fig:img-svm-ap}\includegraphics[width=0.22\textwidth, height=0.15\textheight]{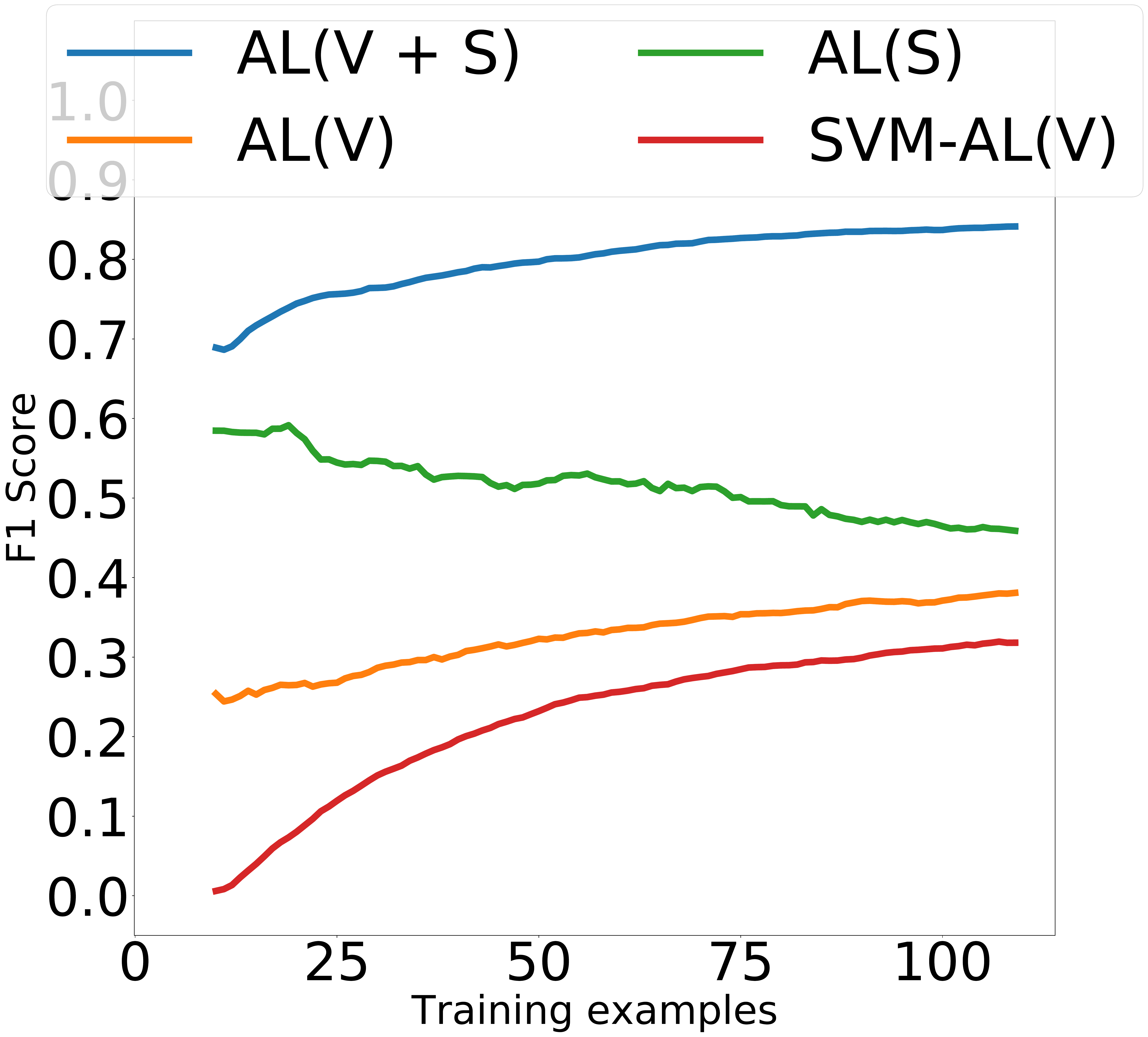}}

\caption{Average Precision Scores for 30 concepts defined on the AWA and Imagenet datasets. We contrast the performance of SVM with visual features against our method  using only visual features, only semantic features and using both visual and semantic features.}
\label{fig:ap-scores}
\end{figure}

\subsection{Effect of different step size $\alpha$}
\begin{figure}[!htb]
	\centering
	\includegraphics[width=0.35\textwidth, height=0.20\textheight]{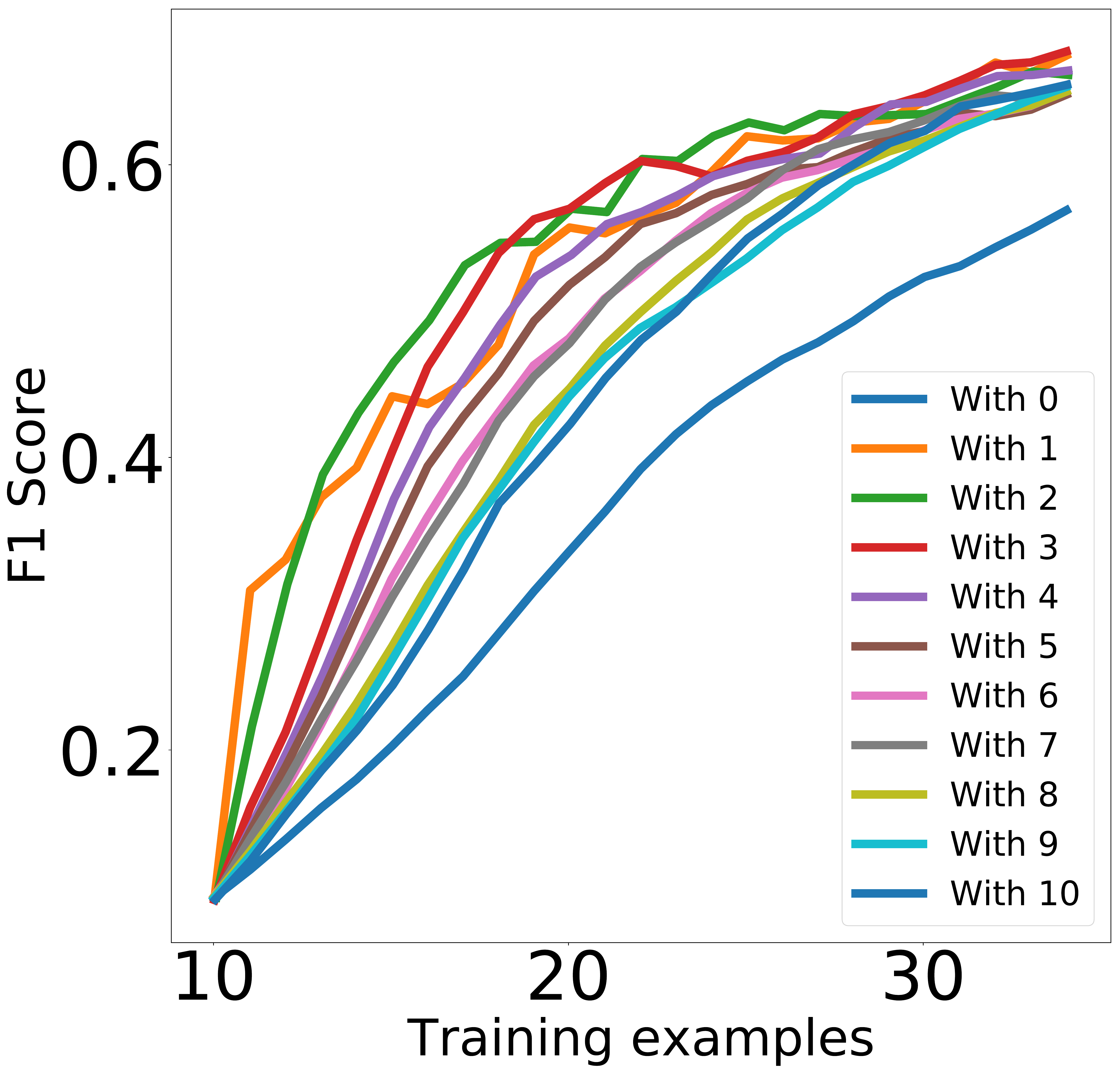}
    \caption{F1 Scores for 30 concepts defined on the AWA dataset. The graph shows the effect of different step sizes.}
    \label{fig:img-ap}
\end{figure}

The figure \ref{fig:img-ap} shows the cross validation results for choosing the step size $\alpha$ in Algorithm 3. This is important since we want to have the step size which is neither too low nor too high. A slow step size will reduce the speed of convergence to the decision boundary of the concept and a high value will lead to unstability in initial runs. To do cross validation we choose 30 concepts different from the ones chosen for evaluation and run our method of adaptive threshold with different values of $\alpha$. Since $\alpha$ = 2 gives the best performance and hence we use this value in our experiments. 

\subsection{F1 scores for individual concepts}
In this section we show the results of F1 scores for individual concepts. We can see our method performs well on all different concepts. The roughness in the curves of Imagenet is attributed to the fact that $f^*$ values changes quite a lot when we try to estimate the ranking of a million points just based on a few points.
\begin{figure*}[!htb]
\foreach \i in {1,...,6}{
     \foreach \j in {1,...,5}{
     \noindent\subfigure{\includegraphics[width=0.19\textwidth, height=0.16\textheight]{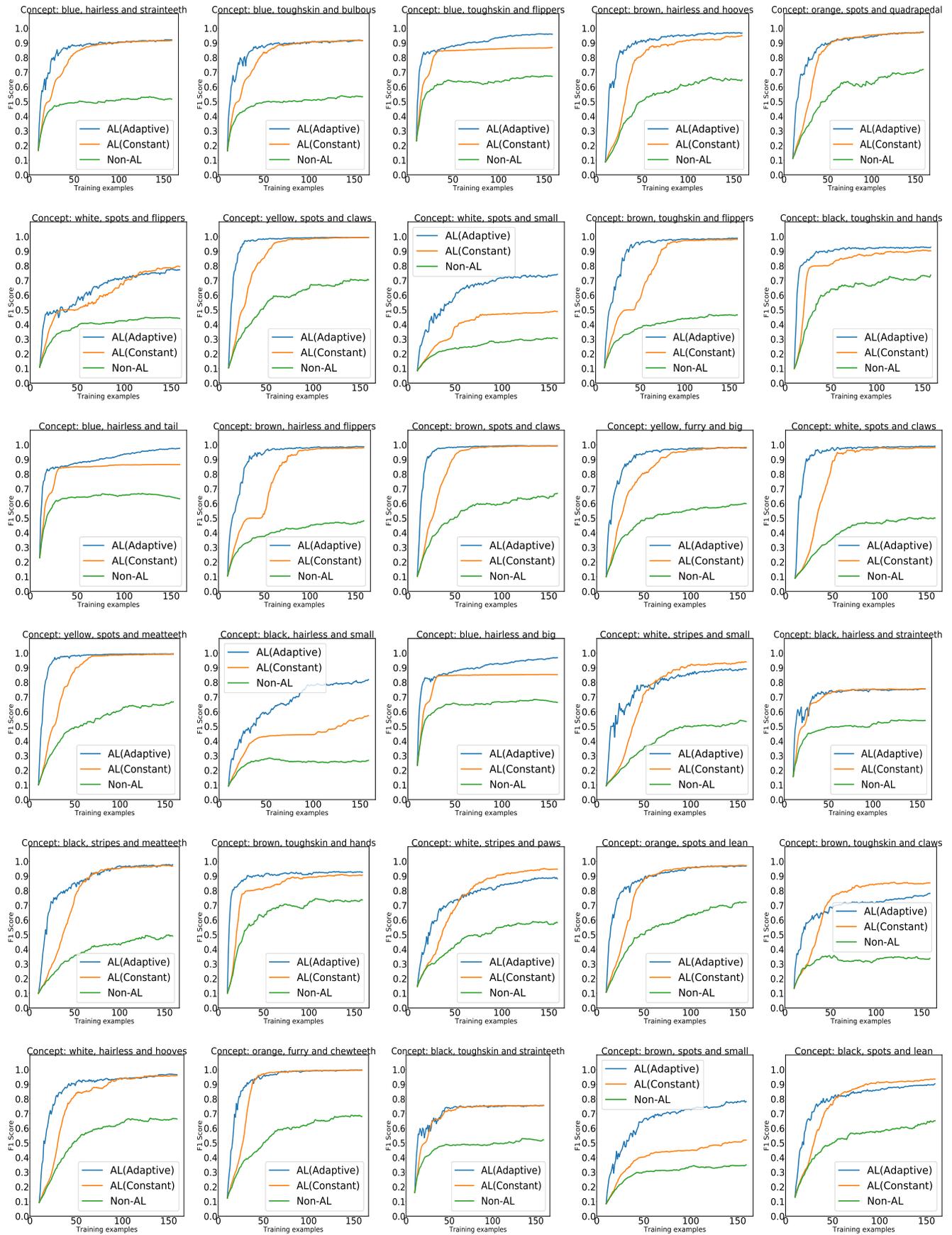}}
   }\\
}
\caption{F1 scores for all concepts defined on AWA dataset}
  \label{fig:f1-awa}
\end{figure*} 

\begin{figure*}

\foreach \i in {1,...,6}{
     \foreach \j in {1,...,5}{
     \subfigure{\includegraphics[width=0.19\textwidth, height=0.16\textheight]{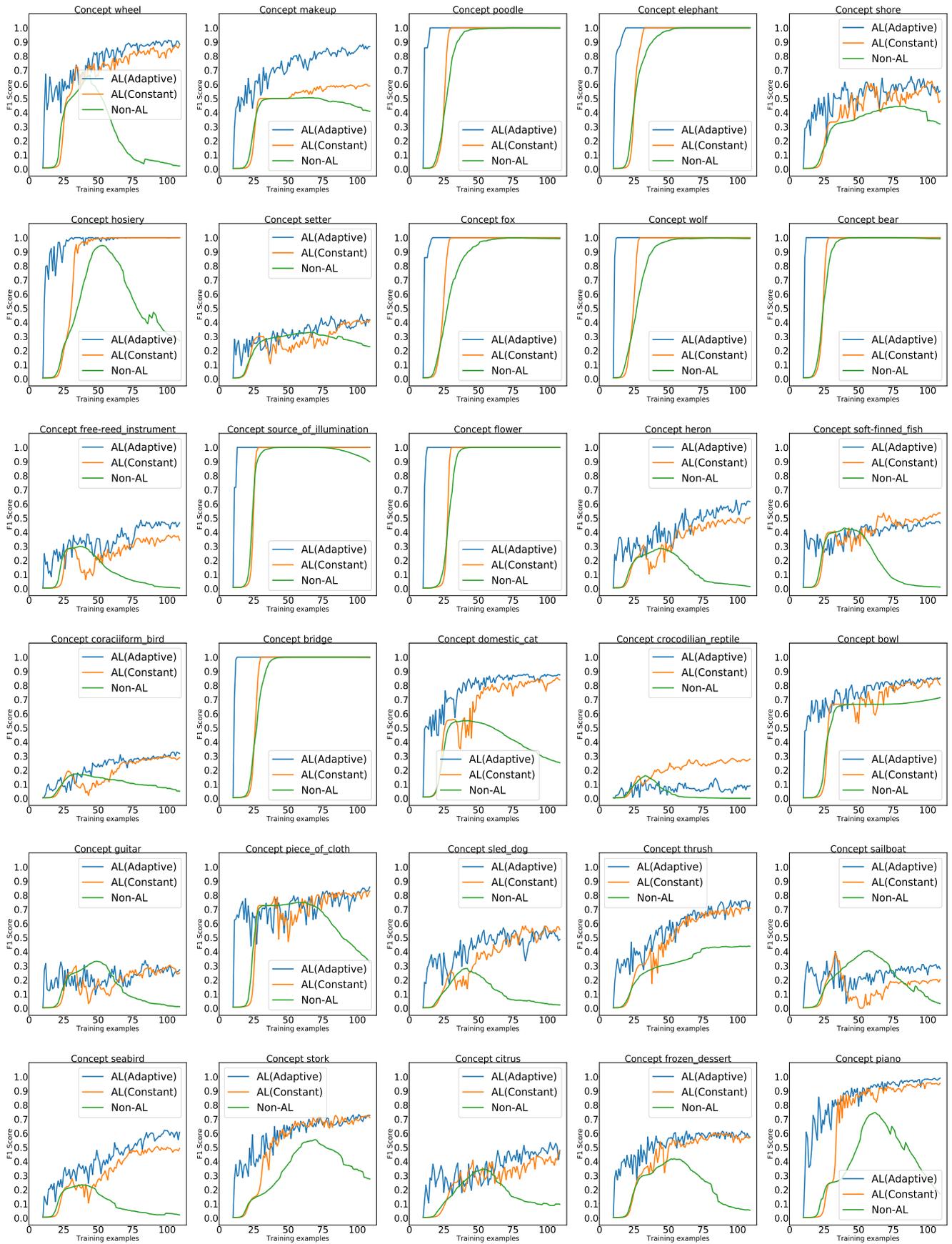}}
   }\\
}
\caption{F1 scores for all concepts defined on Imagenet dataset}
  \label{fig:f1-imagenet}
\end{figure*}

\subsection{Average Precision scores for individual concepts}
In this section we show the results of Average Precision values for individual concepts. We can see our method of adaptive threshold performs well on all different concepts against constant threshold and approach that does not use active learning at all.
\begin{figure*}[!htb]
\foreach \i in {1,...,6}{
     \foreach \j in {1,...,5}{
     \noindent\subfigure{\includegraphics[width=0.18\textwidth, height=0.15\textheight]{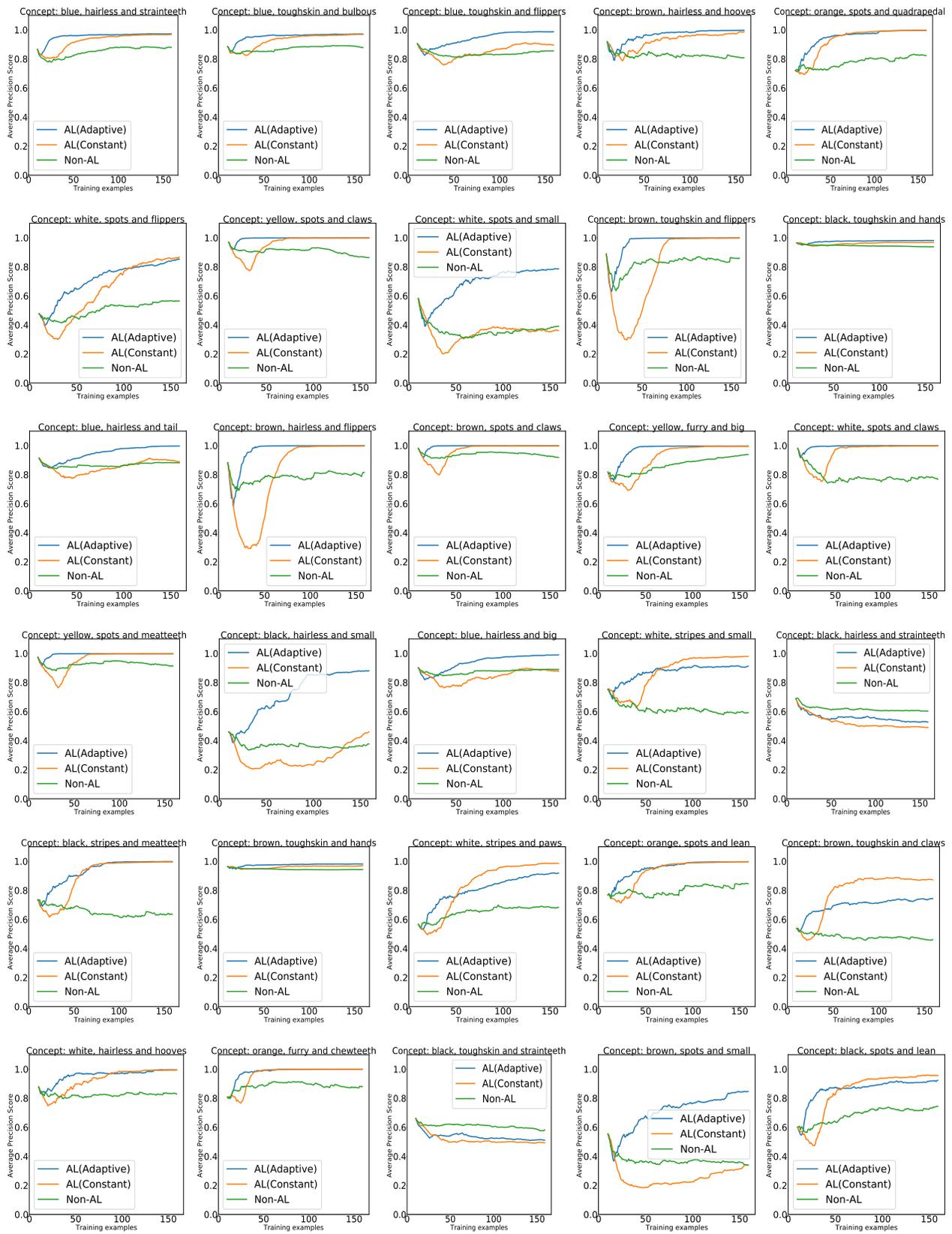}}
   }\\
}
\caption{Average Precision scores for all concepts defined on AWA dataset}
  \label{fig:ap-awa}
\end{figure*}

\begin{figure*}[!htb]
\foreach \i in {1,...,6}{
     \foreach \j in {1,...,5}{
     \noindent\subfigure{\includegraphics[width=0.18\textwidth, height=0.15\textheight]{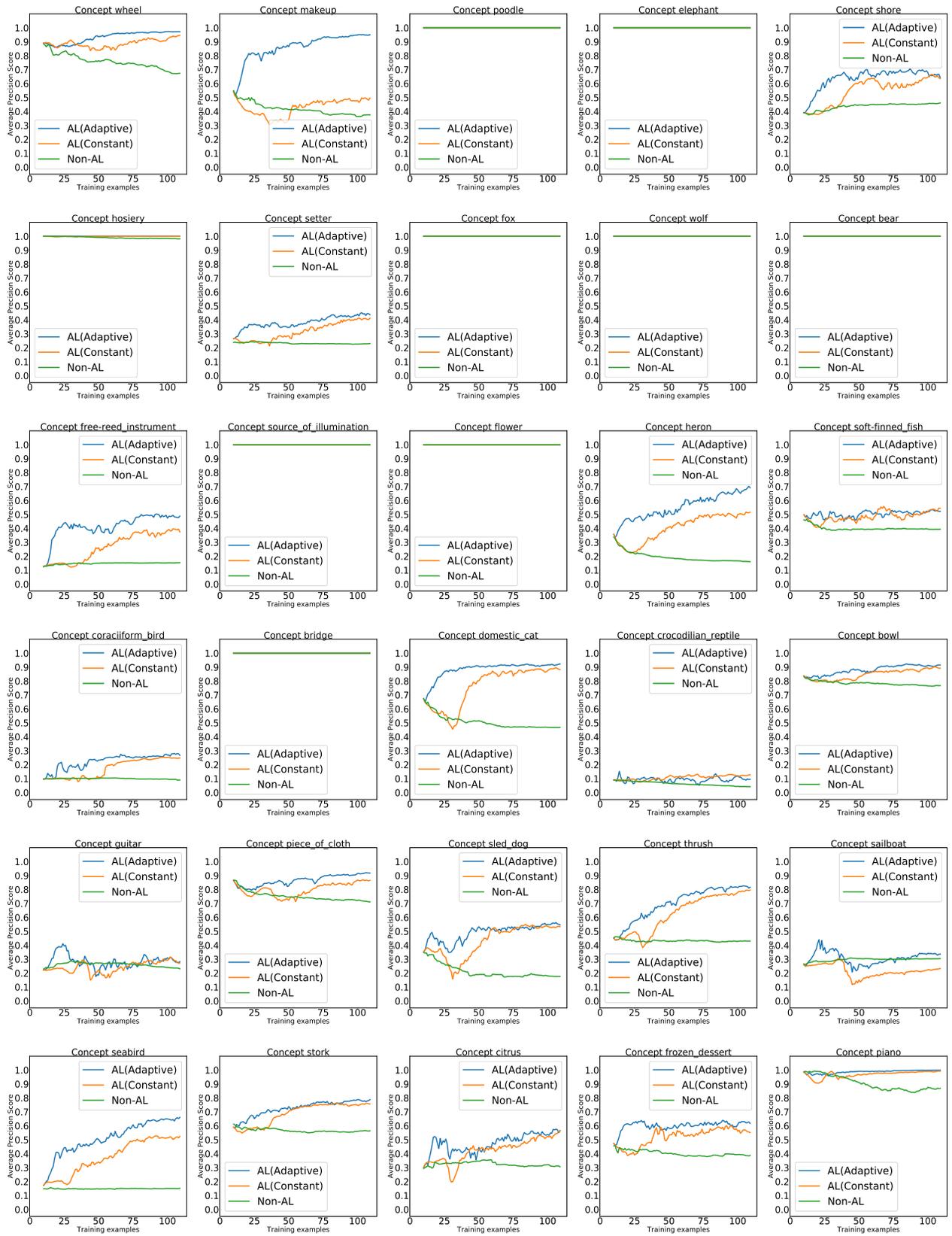}}
   }\\
}
\caption{Average Precision scores for all concepts defined on Imagenet dataset}
  \label{fig:ap-imagenet}
\end{figure*}

\end{document}